\documentclass{article}
\usepackage{iclr2026,times}

\usepackage{amsmath,amsfonts,bm}

\def\eqref#1{equation~\ref{#1}}

\def\1{\bm{1}}

\DeclareMathAlphabet{\mathsfit}{\encodingdefault}{\sfdefault}{m}{sl}
\SetMathAlphabet{\mathsfit}{bold}{\encodingdefault}{\sfdefault}{bx}{n}

\def\gS{{\mathcal{S}}}

\usepackage{booktabs}
\usepackage{hyperref}
\usepackage{url}
\usepackage{amsfonts}
\usepackage{nicefrac}
\usepackage{microtype}
\usepackage{xcolor,colortbl}
\usepackage{arydshln}
\usepackage{color}
\usepackage[normalem]{ulem}
\usepackage{graphicx}
\usepackage{amsmath}
\usepackage{amssymb}
\usepackage{epsfig}
\usepackage{algorithm}  
\usepackage{algpseudocode}
\usepackage{pifont}
\usepackage{multirow,multicol,makecell}
\usepackage{bbm}
\usepackage{mathtools}
\usepackage{diagbox}
\usepackage{enumitem}
\usepackage{soul}
\usepackage{tikz}
\usepackage{bbding}
\usepackage{wrapfig}
\usepackage{colortbl}
\usepackage{amsmath}
\usepackage{algorithm}
\usepackage{xspace}
\usepackage{hhline}
\usepackage{caption}
\usepackage{amsmath}
\usepackage{mathtools}

\definecolor{mygray}{gray}{.9}
\definecolor{mycitecolor}{RGB}{0,101,177}
\hypersetup{colorlinks=true, citecolor=mycitecolor, pdfborder={0 0 0}}
\newcommand{\ie}{\textit{i}.\textit{e}.}

\newcommand{\defe}{\triangleq}

\newcommand{\dist}{\textrm{d}} 
\def\Ours{\textbf{URSA}\xspace}
\def\OursRegular{URSA\xspace} 

\title{Uniform Discrete Diffusion with Metric Path for Video Generation}

\author{
Haoge Deng\textsuperscript{$1,\!3,\!5\!$\,}\thanks{Equal Contribution. 
This work was done when H.Deng, T.Pan, and Y.Liu were interns at BAAI. \\
\hspace*{1.4em}\textsuperscript{\dag}Corresponding Author: 
\textit{wangxinlong@baai.ac.cn},\, \textit{zhaoxiang.zhang@ia.ac.cn}} \,,
Ting Pan\textsuperscript{$2,\!3,\!5*$}, 
Fan Zhang\textsuperscript{$5*$}, 
Yang Liu\textsuperscript{$4,\!5*$}, 
Zhuoyan Luo\textsuperscript{$5$}, 
Yufeng Cui\textsuperscript{$5$} \\
{\hspace{0.25em}\textbf{Wenxuan Wang\textsuperscript{$5$},
Chunhua Shen\textsuperscript{$4$}, 
Shiguang Shan\textsuperscript{$2,\!3$}, 
Zhaoxiang Zhang\textsuperscript{$1,\!3\dag$},
Xinlong Wang\textsuperscript{$5$\dag}}} \\ [3pt]
{\textsuperscript{$1$}National Laboratory of Pattern Recognition, CASIA} \\
{\textsuperscript{$2$}Key Laboratory of Intelligent Information Processing, ICT, CAS} \\
{\textsuperscript{$3$}University of Chinese Academy of Sciences} \\
{\textsuperscript{$4$}Zhejiang University \quad \textsuperscript{$5$}Beijing Academy of Artificial Intelligence}
}

\iclrfinalcopy
\begin{document}

\makeatletter
\let\@oldmaketitle\@maketitle
\renewcommand{\@maketitle}{
\@oldmaketitle
\vspace{-2em}
\centering
\includegraphics[width=.99\linewidth]{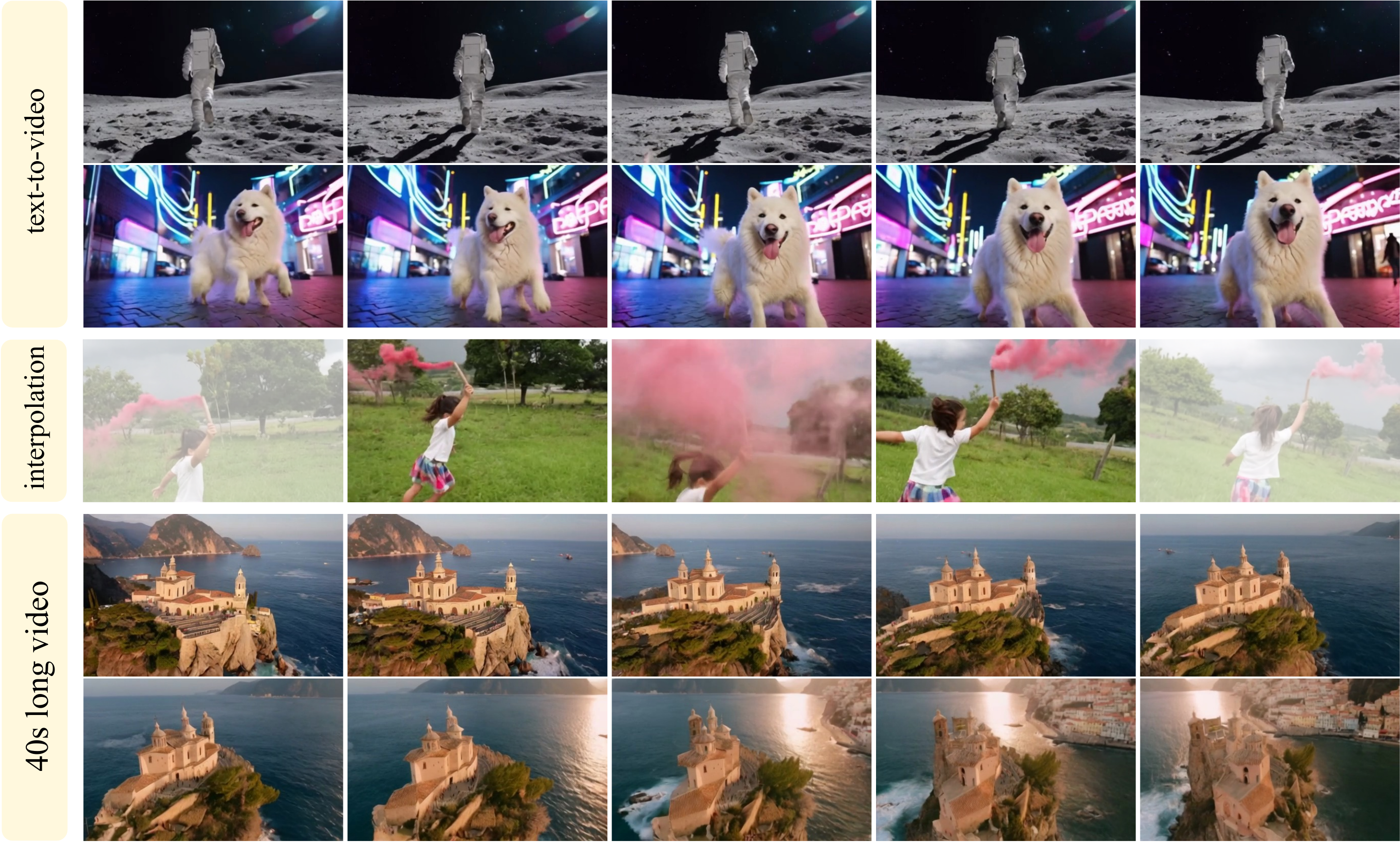} 
\captionsetup{hypcap=false}
\captionof{figure}{
Visualization of \Ours across diverse video generation tasks: text-to-video generation, video interpolation, and long video generation. 
These examples underscore the versatility of \OursRegular.
}\label{fig:teaser}
\vspace{0.5em}
}
\makeatother

\maketitle

\begin{abstract}
Continuous-space video generation has advanced rapidly, 
while discrete approaches lag behind due to error accumulation and long-context inconsistency.
In this work, we revisit discrete generative modeling and present 
\textbf{U}niform disc\textbf{R}ete diffu\textbf{S}ion with metric p\textbf{A}th (\Ours),
a simple yet powerful framework that bridges the gap with continuous approaches for the scalable video generation.
At its core, \OursRegular formulates the video generation task as an iterative global refinement of discrete spatiotemporal tokens.
It integrates two key designs: 
a Linearized Metric Path and a Resolution-dependent Timestep Shifting mechanism.
These designs enable \OursRegular to scale efficiently to high-resolution image synthesis and long-duration video generation, while requiring significantly fewer inference steps.
Additionally, we introduce an asynchronous temporal fine-tuning strategy that unifies versatile tasks within a single model, 
including interpolation and image-to-video generation.
Extensive experiments on challenging video and image generation benchmarks demonstrate that \OursRegular consistently outperforms existing discrete methods and achieves performance comparable to state-of-the-art continuous diffusion methods.
Code and models are available at {\small \url{https://github.com/baaivision/URSA}}
\end{abstract}

\section{Introduction}

Continuous-space visual generation has achieved remarkable progress in both image and video synthesis~\citep{MODEL:FLUX,MODEL:IMAGEN3,MODEL:DALLE3,MODEL:SORA,MODEL:WAN,MODEL:SEEDANCE,MODEL:COGVIDEOX,MODEL:HUNYUANVIDEO}.
Driven by advances in diffusion model algorithms~\citep{ALGO:DDPM,ALGO:SMLM}, these continuous-space methods have demonstrated strong capabilities in producing high-fidelity and visually coherent content, establishing themselves as the dominant paradigm for generative modeling.

In parallel, discrete-space text generation has become the \textit{de facto} paradigm for large language models~\citep{LLM:GPT1,LLM:GPT2,LLM:GPT3}.
Inspired by the success of LLMs, recent works have extended similar ideas to visual generation through discrete tokenization, using either next-token prediction~\citep{MODEL:LLAMAGEN,VLM:EMU3, MODEL:VIDEOPOET} or masked token prediction~\citep{MODEL:MUSE,MODEL:SHOWO}.
However, discrete approaches still lag behind their continuous counterparts, facing challenges such as error accumulation and maintaining long-context consistency, especially in video generation.
For instance, even though masked diffusion models employ bidirectional transformers, we still observe low visual quality and unnatural object motions.

\begin{figure*}[t]
\centering
\includegraphics[width=0.95\linewidth]{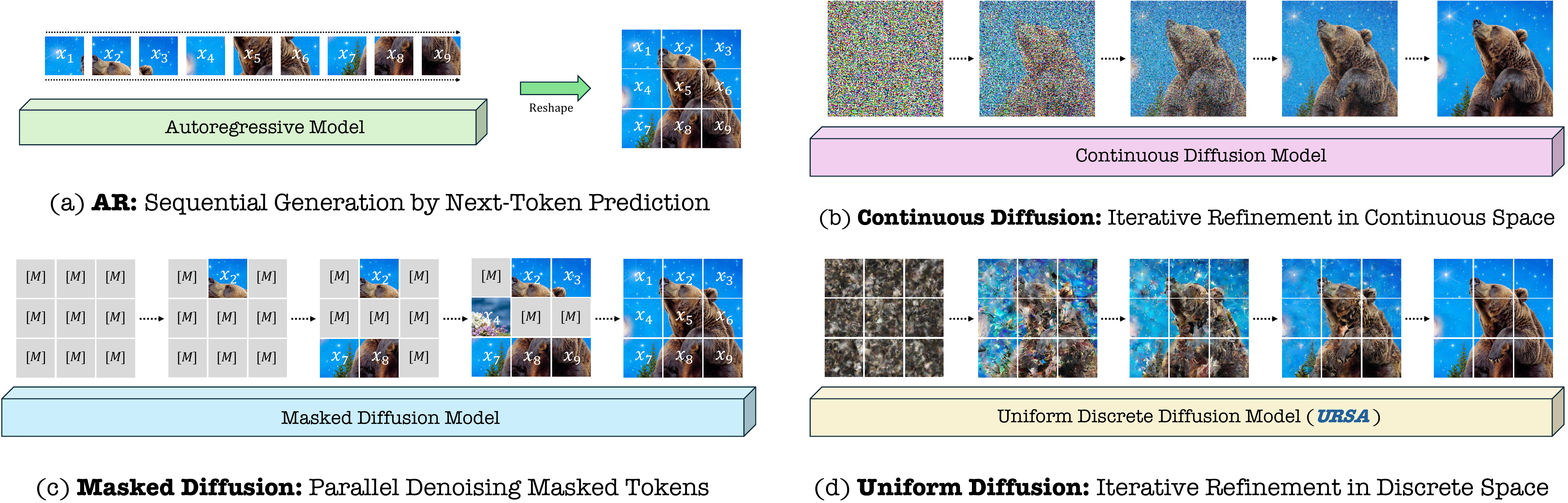}
\caption{
\textbf{Illustration of different image/video generation paradigms.} Discrete-space approaches such as AR and MDM adopt non-refinable local generation, where produced tokens are fixed once generated. In contrast, \OursRegular introduces iterative global refinement, conceptually aligning discrete methods with continuous-space approaches, and substantially narrowing their performance gap.
}
\label{fig:illustration}
\end{figure*}

In this work, we first revisit discrete generative modeling and introduce \Ours, a powerful visual generation framework built upon \textbf{U}niform disc\textbf{R}ete diffu\textbf{S}ion 
with metric p\textbf{A}th. 
Our approach is simple: we generate videos and images by iterative refinement over discrete spatiotemporal tokens.
As illustrated in Fig.~\ref{fig:illustration}, unlike classic autoregressive (AR) models and masked diffusion models (MDM) that adopt non-refinable local generation, where produced tokens are fixed once generated, \OursRegular emphasizes \textit{iterative refinement over global discrete tokens}, conceptually aligning discrete methods with continuous counterparts, and substantially narrowing their performance gap.
\OursRegular starts from categorical noise, $x_0 \sim \mathrm{Unif}([K])^D$, where each of the $D$-dimensional discrete token is independently sampled from a uniform distribution over the vocabulary $[K] = \{1,2,\dots, K\}$, and iteratively performs global refinement along a metric-guided probability path to obtain $x_1$ on the data manifold, \ie, the target image or video. This iterative process enables \OursRegular to capture the hierarchical structure of video data, from global layouts to detailed dynamics, while leveraging temporal redundancy to preserve spatiotemporal coherence. 

We propose a novel metric probability path tailored for long visual sequences by incorporating two key components: a linearized metric path and a resolution-dependent timestep shifting mechanism. Collectively, these designs enable precise control over data perturbations, a property essential for effectively learning hierarchical data manifolds.
This construction allows \OursRegular to scale efficiently to long-sequence tasks, such as high-resolution image synthesis and long video generation, while requiring substantially fewer inference steps.
Furthermore, we introduce an asynchronous timestep scheduling strategy, where timesteps are independently sampled for each frame. This asynchronous design empowers \OursRegular to generate minute-level long videos and support a wide range of tasks within a unified model, including image-to-video generation, video interpolation, and extrapolation.

\OursRegular achieves a text-to-video score of 82.4 on VBench~\citep{EVAL:VBENCH}, outperforming discrete and continuous baselines. 
In image-to-video generation tasks, \OursRegular reaches a VBench score of 86.2, on par with the state-of-the-art open-source models. 
For text-to-image generation, \OursRegular achieves a DPG-Bench~\citep{EVAL:DPGBENCH} score of 86.0, exceeding previous discrete approaches. 
Moreover, \OursRegular exhibits strong zero-shot generalization in variable length contexts, underscoring its versatility.

Our contributions can be summarized as follows:
1) We propose \Ours, a simple yet powerful framework that bridges the gap to continuous diffusion methods and enables scalable video generation. 
2) We highlight two key designs, Linearized Metric Path and Resolution-dependent Timestep Shifting for long-sequence training. We further propose an asynchronous timestep scheduling strategy that enables multi-task video generation.
3) \OursRegular substantially pushes the envelope of discrete generation, attaining state-of-the-art performance on VBench, DPG-Bench, and GenEval~\citep{EVAL:GENEVAL}.

\section{Related Works}

\subsection{Continuous-Space Visual Generation}

Continuous methods for visual generation have achieved significant progress in recent years.
Early endeavors such as variational autoencoders (VAEs)~\citep{TOKENIZER:VAE} and flow-based models~\citep{MODEL:NICE,MODEL:REALNVP} exploit continuous latent spaces to model complex images, while GANs~\citep{ALGO:GAN} generate high-resolution images with strong perceptual quality via adversarial training~\citep{MODEL:BIGGAN,MODEL:STYLEGAN}.
Diffusion models~\citep{ALGO:DDPM,ALGO:SMLM}, which learn to recover data by progressively denoising Gaussian noise in a continuous space, demonstrated remarkable performance in both image and video generation~\citep{MODEL:SEEDREAM,MODEL:FLUX,MODEL:IMAGEN3,MODEL:DALLE3,MODEL:QWENIMAGE,MODEL:SORA,MODEL:HUNYUANVIDEO,MODEL:SEEDANCE,MODEL:WAN,MODEL:KLING,MODEL:STEPVIDEO}.
MAR~\citep{MODEL:MAR} employs an autoregressive framework with a diffusion head to produce continuous-valued outputs, and NOVA~\citep{MODEL:NOVA} further extends this idea to video generation, applying autoregressive modeling to spatiotemporal sequences.
\OursRegular shares the same spirit as continuous diffusion models, performing global iterative refinement, but operates over discrete tokens.

\subsection{Discrete-Space Visual Generation}

Discrete visual generation can be broadly categorized into autoregressive and masked diffusion models, both operating on discrete visual tokens such as pixels~\citep{MODEL:VIDEOPIXEL,MODEL:PARALLEL} or latent codes~\citep{TOKENIZER:VQVAE,TOKENIZER:VQGAN}.
Autoregressive models generate discrete visual tokens sequentially, with each prediction conditioned on previously generated context. This approach has been applied to both image~\citep{MODEL:LLAMAGEN,MODEL:DALLE1,MODEL:COGVIEW1,MODEL:PARTI, ALGO:Dimo} and video synthesis~\citep{VLM:EMU3,MODEL:VIDEOGPT,MODEL:VIDEOPOET,MODEL:LOONG}. Although simple in concept, this design often has slow inference and significant error accumulation.
In contrast to autoregressive methods, masked diffusion models~\citep{ALGO:DFM,MODEL:MASKGIT,MODEL:MUSE,MODEL:MAGVIT} introduce the prediction of masked tokens, enabling parallel generation and improved modeling of global context. Despite these advantages, it remains challenging to apply these methods to long sequences, \textit{e.g.} high-fidelity long-form video. 
FUDOKI~\citep{MODEL:FUDOKI} investigates the integration of discrete flow matching~\citep{ALGO:DFM} within native multimodal models.
In this work, we adopt a uniform discrete diffusion approach, which performs iterative global refinement from categorical noise. By addressing key challenges, \OursRegular enables both efficient inference and high-quality long-sequence generation.

\section{Methodology}

We first review the concepts of uniform discrete diffusion / discrete flow matching in Sec.~\ref{sec:pre}, which provide the theoretical foundation for our framework. In Sec.~\ref{sec:path}-\ref{sec:fips}, we introduce \OursRegular, a simple yet powerful framework that bridges the gap between discrete and continuous approaches, enabling effective and scalable video generation.

\subsection{Preliminary: Discrete Flow Matching}\label{sec:pre}

Discrete Flow Matching (DFM)~\citep{ALGO:DFM,ALGO:KINETIC} introduces a family of generative models designed to map data from an initial distribution $p_0(x)$, to a final distribution $p_1(x)$, within a discrete state space. The model utilizes a time-dependent probability path, $p_t(x)$, which interpolates between these two distributions over the interval $t \in [0, 1]$. The key idea behind DFM is to define a velocity field, $u_t$, which drives the evolution of this probability path, enabling the model to simulate a Markov process and generate new data samples.

\textbf{Probability paths.}
We consider the probability path $p_t(x)$, where $t$~$\in$~$[0,1]$ indexes a time-dependent probability distribution
between a source distribution $p_0(x)$ and a target distribution $p_1(x)$ over $t$. 
Given a data distribution $q(x)$ over $x=(x^1, \ldots, x^D) \in [K]^D$, the probability path is defined as
\begin{align}\label{eq:prob_path}
p_t(x) \defe \sum_{x_1 \in \gS} p_t(x | x_1) q(x_1), \text{ where } p_t(x|x_1) \defe \prod_{i=1}^D p_t(x^i|x_1^i),     
\end{align}

$p_t(x^i \mid x_1^i)$ denotes a \emph{conditional} forward probability path, characterizing the evolution of the state $x^i$ given the initial state $x_1^i$. 

\textbf{Probability velocities.}
To generate the predefined probability path $p_{t}(x)$, we consider a Continuous-Time Markov Chain (CTMC), modeled as a stochastic process. The dynamics of this CTMC are governed by a probability velocity $u_t$, also known as the  \emph{transition rate}. The transition rate models how the current state $x_t$ evolves toward the target state $x_1$ over time. Within this framework, each token $i$ is updated independently according to the following transition rule:
\begin{equation}\label{eq:sample}
x_{t+h}^i \sim \delta_{x_t^i}(\cdot) + h \, u_t^i(\cdot \mid x_t^i, x_1^i),
\end{equation}
where $u_t^i(\cdot \mid x_t^i, x_1^i)$ represents \emph{velocity field}, a conditional rate function that governs the flow of probability from the current state $x_t^i$ to the target state $x_1^i$ over time.
Equation (\ref{eq:sample}) can be interpreted as a small perturbation of the point mass $\delta_{x_t^i}$, scaled by the step size $h$, effectively modeling discrete state transitions as a continuous-time stochastic process. This velocity field is central to DFM, as it characterizes the dynamics of the probability path and is the primary quantity learned during training. 

\subsection{Uniform Discrete Diffusion with Metric Path}\label{sec:uddm}

We present \OursRegular, a novel framework built on uniform discrete diffusion with a metric path for image and video generation. 
In this section, we first introduce three key innovations:
(1) a Linearized Metric-Path for structured and tractable trajectory design, (2) a Resolution-dependent
Timestep Shifting mechanism to improve training stability and representation learning for long
video sequences, and (3) a Frame-wise Independent
Perturbation Scheduling strategy for unified long-video generation and multitask learning.
After introducing these core components, we further provide the training procedure and sampling algorithm.

\begin{figure}[t]
\centering 
\vspace{-3mm}
\includegraphics[width=0.98\linewidth,trim= 0 0 0 0,clip]{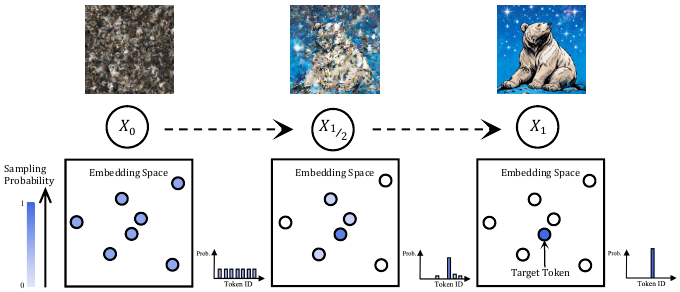} 
\caption{\textbf{Global refinement via token distance in embedding space.} Starting from categorical noise $x_0$ (left), our framework refines data based on token distance to get target data $x_1$ (right), enabling hierarchical structure generation from global semantics to fine details.}
\vspace{-3mm}
\label{fig:fig2_refinement}
\end{figure}

\subsubsection{Metric Probability Path for Long Sequence Data}\label{sec:path}

For data with varying sequence lengths, the degree of perturbation should be adapted during training. This requires a probability path to effectively handle sequences of different lengths, such as high-resolution images or videos. In this section, we introduce two key techniques, linearized metric path and resolution-dependent timestep shifting, to address this challenge, ensuring that the perturbation process is appropriately adjusted based on the sequence length.

\textbf{Linearized metric path.} Inspired by ~\citep{ALGO:KINETIC}, We introduce the linearized metric path, a novel probability path derived from token embedding distances. Formally, we define the distance function
$\dist: \mathcal{T} \times \mathcal{T} \to \mathbb{R}_{\geq 0}$,
which measures the discrepancy between the codebook embeddings of generated token $x$ and the target tokens $x_1$. 
The distance satisfies the property $\dist(x,x_1)=0\Leftrightarrow x=x_1$, ensuring a well-defined metric structure. 
Based on this, the probability path is defined as
\begin{align}\label{eq:metric_prob_path}
p_t(x | x_1) =  \text{softmax}\left( -\beta_t \dist(x, x_1) \right) = \prod_{i=0} \text{softmax}\left( -\beta_t \dist(x^i, x_1^i) \right),
\end{align}
where $\beta_t:[0,1] \to \mathbb{R}_{\ge 0}$ is a monotonic scheduler function with boundary conditions $\beta_0 = 0$, $\beta_1 = \infty$.
The core of linearized path lies in the functional form of $\beta_t$, which is parameterized as
\begin{align}\label{eq:beta_t}
\beta_t =  c \times (\dfrac{t}{1-t})^\alpha, t \in [0,1),
\end{align}
where $c > 0$ and $\alpha > 0$ are hyperparameters that control the relationship between the sampling distance $\dist(x_t, x_1)$ and time $t$. Specifically, the forward process samples $x_t \sim p_{t \mid 1}(\cdot \mid x_1)$, with boundary conditions yielding a uniform distribution over codebook embeddings at $t = 0$ and a deterministic sample at $x_1$ when $t = 1$, illustrated in Figure~\ref{fig:fig2_refinement}.
When $t$ is between $0$ and $1$, our objective is to find an appropriate set of values for $c$ and $\alpha$ that preserve the linear relationship between $t$ and $\dist(x_t, x_1)$. This linearity provides a finer control of perturbations over the probability path, as described next. 
Additional experiments and discussions on the impact of linearized metric path on model convergence speed and overall performance are provided in Sec.~\ref{sec:ablation}.

\textbf{Resolution-dependent timestep shifting.}
Intuitively, since higher resolutions contain more pixels, more perturbation is needed to alter the signal. To address this, we introduce a time shift parameter $\lambda$, which adjusts the timestep based on the resolution. For any given $t$, we define the shifted timestep $\tilde{t}$
\begin{align}\label{eq:timestep_shift}
\tilde{t} = \dfrac{t}{t + \lambda(1-t)}.
\end{align}
Because our proposed linearized metric path enforces a linear relationship between $t$ and $\dist(x_t, x_1)$, we modulate this path using $\lambda$ to accommodate varying data resolutions. For higher resolutions, we set $\lambda>1$ to create a convex relationship between $\tilde{t}$ and $\dist(x_t, x_1)$ that introduces stronger perturbations.
For lower resolutions, we set $\lambda < 1$, yielding a concave relationship with more gradual perturbations.

\subsubsection{Asynchronous Timestep Scheduling}\label{sec:fips}

Due to their complex spatiotemporal dynamics and broad applicability across downstream tasks, prior video generation methods render task-specific modeling both inefficient and resource-intensive.
Motivated by diffusion forcing~\citep{ALGO:DForcing}, we propose an asynchronous timestep scheduling strategy tailored for multi-task training and sampling. Rather than applying the same noise level across all frames in a video sequence (\ie, synchronous timestep scheduling), we assign uniform noise levels independently to each frame.
Formally, given a clean video sequence $\mathbf{F} = \{f^{(1)}, f^{(2)}, \dots, f^{(n)}\}$ with $n$ frames, we assign each frame a continuous time $t_i \sim \mathcal{U}(0,1)$, forming the timestep schedule $\mathbf{T} = \{t_1, t_2, \dots, t_n\}$. The corresponding noisy sequence is denoted as $\tilde{\mathbf{F}} = \{f^{(1)}_{t_1}, f^{(2)}_{t_2}, \dots, f^{(n)}_{t_n}\}$, where the metric-induced probability path in Eq.~\ref{eq:metric_prob_path} is applied frame-wise according to its assigned $t_i$. \\
This strategy enables fine-grained temporal modeling over the timestep schedule via decoupling noise levels across frames. As a result, the training progress adaptively balances local frame reconstruction with global temporal coherence, facilitating versatile generation objectives, such as text-to-video, image-to-video, video extrapolation, and start–end frame control within a unified model architecture. 
Additional visualizations for these advanced generation tasks are provided in Appendix~\ref{sec:supp:extrapolation}~\&~\ref{sec:supp:startend}.

\subsubsection{Training and Sampling}

\textbf{Training.} We first encode video clips into discrete token sequences using a pre-trained tokenizer, resulting in a clean video sequence $x_1=\{f_1^{(1)}, f_1^{(2)}, \dots, f_1^{(n)}\}$, where $n$ denotes the number of video frames and $f_1^{(i)}$ denotes the $i$-th frame tokens. At each training step, we uniformly sample timesteps $t_{i} \in [0, 1]$ for each frame $f_1^{(i)}$ in the sequence and obtain a perturbed sequence $x_t \sim p_{t}(\cdot \mid x_1)$ via the proposed metric probability path. The backbone, implemented with the LLM architecture, takes as input the concatenation of text tokens $e$ and noisy tokens $x_t$, and produces logits over the token vocabulary to predict the original sequence $x_1$.
The training objective is formulated as the expected cross-entropy between the ground-truth visual tokens and the model’s predicted distribution:
\begin{equation}
\label{eq:training_loss}
\mathcal{L} = \mathbb{E}_{t \sim \mathcal{U}[0,1],\, {x}_1, {x}_t}
\left[ -\log p_{1 \mid t}\left( {x}_1 \mid {x}_t, {e}\right) \right].
\end{equation}

\textbf{Sampling.} We follow \citet{ALGO:DFM, ALGO:KINETIC} and employ the Euler solver for efficient and high-quality generation. Specifically, we first uniformly sample $x_0$ from the full vision vocabulary and feed it into the model to obtain the prediction $\hat{x}_1$. Following \citet{ALGO:KINETIC}, we compute the velocity field $u_t(\cdot \mid x_t, \hat{x}_1)$. We then iteratively refine $x_t$ using $u_t$, where each iteration updates the sample $x_t$ along the estimated direction. Once the $T$ refinement steps have been completed, the sampling process returns a clean image or video. Additional details are provided in Appendix~\ref{sec:supp:training_sampling}.

\section{Experiment}

\subsection{Experiment Setup}\label{sec:experiment:setup}

\textbf{Datasets.}
We leverage a curated selection of high-quality datasets to effectively train \OursRegular models.
For text-to-image training, we collect 16M image-text pairs sourced from Unsplash~\citep{DATASET:UNSPLASH}, DataComp~\citep{DATASET:DATACOMP}, COYO~\citep{DATASET:COYO}, and JourneyDB~\citep{DATASET:JOURNEYDB}.
These pairs are filtered by image resolution and aesthetic score, and further supplemented with 14M AI-generated image samples using the FLUX.1 model~\citep{MODEL:FLUX}.
For text-to-video training, we select 12M video-text pairs from the highest scoring subset of Koala-36M~\citep{DATASET:KOALA} and complement them with 12M internal video-text pairs.
The internal videos are captioned using the Emu2-17B model~\citep{VLM:EMU2} in conjunction with the captioning engine~\citep{VLM:EVE}.
We uniformly sample short and long captions during training, with a maximum length of 320 tokens.

\textbf{Architectures.}
We initialize our visual generation model with weights from a pre-trained LLM.
Specifically, we adopt the Qwen3 LLM architecture~\citep{LLM:QWEN3}, which natively incorporates QK-Norm~\citep{MODEL:VIT22B} layer to stabilize the multimodal training.
To better capture the spatiotemporal structure inherent in videos, we introduce an enhanced M-RoPE~\citep{VLM:QWEN2VL} that allocates interleaved frequency components across temporal, height, and width dimensions, following the approach of Mogao~\citep{MODEL:MOGAO}.
Crucially, unlike~\citet{MODEL:MOGAO}, our 3D-RoPE assigns identical positions for texts, ensuring equivalence with the 1D-RoPE~\citep{ARCH:ROPE}.
We use the Cosmos~\citep{MODEL:COSMOS} tokenizer to extract image and video tokens, achieving 4$\times$ temporal and 8$\times$8 spatial compression through a 64K FSQ~\citep{TOKENIZER:FSQ} codebook.
Furthermore, we train an IBQ~\citep{TOKENIZER:IBQ} tokenizer for high-resolution image generation, facilitating efficient 16$\times$16 spatial compression via a 256-dimensional codebook with 131K entries.

\textbf{Diffusion schedulers.}
We adopt the Kinetic Optimal Scheduler~\citep{ALGO:KINETIC}, equipped with a metric-induced probability path specifically designed for the embedding space of vision tokenizers.
Following~\citet{ALGO:KINETIC}, we perform a grid search over the path hyperparameters $\alpha$ and $c$, visually inspecting the reconstructed samples for each $(\alpha,c)$ that fully exploit the time interval $[0,1]$.
Eventually, we select $(\alpha,c)$ to $(1.0,5)$ for the Cosmos tokenizer and $(0.5,6)$ for our IBQ tokenizer. 
For standard uniform diffusion, we use the mixture probability path proposed by~\citet{ALGO:DFM}.
In contrast, for masked diffusion, we adopt the MaskGIT~\citep{MODEL:MASKGIT} scheduler, 
which has been empirically shown to achieve state-of-the-art performance in both image and video generation models~\citep{MODEL:VIDEOPOET,MODEL:MEISSONIC}.
Following established practice in continuous diffusion models, we default to 25 inference steps for image generation and 50 for video generation.

\textbf{Training details.}
\OursRegular is trained on 128 A100 (40GB) GPUs.
In all experiments, we use the AdamW optimizer~\citep{OPTIM:ADAMW} with $\beta_{1}=0.9$, $\beta_{2}=0.95$, weight decay of 0.05, and an initial learning rate of 1e-4. The learning rate employs cosine decay~\citep{OPTIM:SGDR}.
We first pre-train text-to-image models and leverage their weights to initialize text-to-video models.
Subsequently, following \citet{MODEL:SKYREELSV2}, we adapt full-sequence video diffusion models to diffusion forcing architectures by applying frame-wise noise schedules for autoregressive generation.

\textbf{Evaluation.}
We evaluate text-to-image alignment using benchmarks DPG-Bench~\citep{EVAL:DPGBENCH} and GenEval~\citep{EVAL:GENEVAL}.
Each image is generated from original or rewritten text prompts, with resolution determined by model type: 1024$\times$1024 for image generation models to support high fidelity, and 512$\times$320 for video generation models to effectively measure cross-modal generalization.
We access text-to-video generation using VBench~\citep{EVAL:VBENCH} and image-to-video generation with VBench++~\citep{EVAL:VBENCHPLUS}, its comprehensive successor tailored for real-world scenarios.
The videos, sized 49$\times$512$\times$320, are generated from rewritten prompts for text-to-video evaluation, and from original text prompts with official cropped first-frame images for image-to-video evaluation.
We apply classifier-free guidance~\citep{ALGO:CFG} with a scale value of 7.0 in all evaluations.

\subsection{Main results}\label{sec:experiment:main}

\textbf{\OursRegular rivals Sora-like text-to-video generation models despite using a discrete video tokenizer.}
Current discrete video tokenizers offer limited spatiotemporal compression and reconstruction quality, posing significant challenges to bidirectional diffusion transformers.
However, \OursRegular excels in generating video clips from text, achieving strong performance on the VBench, as shown in Table~\ref{tab:res-t2v}.
Compared to Sora-like diffusion models: Vchitect~\citep{MODEL:VCHITECT2}, Pyramid Flow~\citep{MODEL:PYRAMIDFLOW}, LuminaVideo~\citep{MODEL:LUMINAVIDEO}, OpenSora~\citep{MODEL:OPENSORA} and OpenSoraPlan~\citep{MODEL:OPENSORAPLAN},
\OursRegular matches or exceeds their performance, particularly in the semantic field.
These results further underscore the need for a tokenizer that satisfies the imaging quality of state-of-the-art continuous models~\citep{MODEL:HUNYUANVIDEO,MODEL:MAGI1,MODEL:STEPVIDEO,MODEL:COGVIDEOX,MODEL:WAN}.

\begin{table}[ht!]
\caption{\textbf{Text-to-video evaluation on VBench.}
For clarity and to better highlight distinctions between models, 
we report only the most relevant metrics across the quality and semantic dimensions.
}
\vspace{-3mm}
\label{tab:res-t2v}
\begin{center}
\resizebox{\linewidth}{!}{
\renewcommand\arraystretch{1.1} 
\begin{tabular}{lccccccccccccc}
\toprule
Model & \#params & \#videos & 
\makecell{Total\\Score} & \makecell{Quality\\Score} & \makecell{Semantic\\Score} & 
\makecell{Dynamic\\Degree} & \makecell{Aesthetic\\Quality} & \makecell{Imaging\\Quality} & 
\makecell{Object\\Class} & \makecell{Multiple\\Objects} & 
\makecell{Spatial\\Relationship} & \makecell{Color} & \makecell{Scene} \\
\midrule
\rowcolor{mygray} \multicolumn{14}{l}{$\blacktriangledown$ \emph{Continuous models}}  \\
NOVA              & 0.6B & 20M  & 80.1 & 80.4 & 79.1 & 20.1 & 59.4 & 59.4 & 92.0 & 77.5 & 77.5 & 87.7 & 54.1 \\  
Vchitect-2.0      & 2B   & 134M & 81.6 & 82.5 & 77.8 & 58.3 & 61.5 & 65.6 & 87.8 & 69.4 & 54.6 & 86.9 & 57.5 \\
Pyramid Flow      & 2B   & 10M  & 81.7 & 84.7 & 69.6 & 64.6 & 63.3 & 65.0 & 86.7 & 50.7 & 59.5 & 82.9 & 43.2 \\
LuminaVideo       & 2B   & 12M  & 83.0 & 83.9 & 79.3 & 67.1 & 62.3 & 64.6 & 91.0 & 68.3 & 67.3 & 90.2 & 56.1 \\
OpenSoraPlan v1.5 & 8B  & 40M   & 83.0 & 84.2 & 78.2 & 64.4 & 66.9 & 68.5 & 91.9 & 70.7 & 80.1 & 81.8 & 52.1 \\
OpenSora 2.0      & 11B & 85M   & 83.6 & 84.4 & 80.3 & 56.4 & 65.3 & 65.7 & 94.6 & 78.0 & 76.8 & 86.3 & 53.4 \\
MAGI-1            & 24B  & -    & 81.8 & 84.7 & 70.4 & 72.5 & 59.3 & 65.3 & 84.1 & 50.6 & 73.0 & 87.5 & 28.9 \\
Step-Video        & 30B  & -    & 81.8 & 84.5 & 71.3 & 53.1 & 61.2 & 70.6 & 80.6 & 50.6 & 71.5 & 88.3 & 24.4 \\
CogVideoX1.5      & 5B   & -    & 82.0 & 82.7 & 79.2 & 56.2 & 62.1 & 65.3 & 83.4 & 65.3 & 79.4 & 88.4 & 53.3 \\
HunyuanVideo      & 13B  & -    & 83.2 & 85.1 & 75.8 & 70.8 & 60.4 & 67.6 & 86.1 & 68.6 & 68.7 & 91.6 & 53.9 \\
Wan2.1            & 14B  & -    & 83.7 & 85.6 & 76.1 & 65.5 & 66.1 & 69.4 & 86.3 & 69.6 & 75.4 & 88.6 & 45.8 \\
\rowcolor{mygray} \multicolumn{14}{l}{$\blacktriangledown$ \emph{Discrete models}}                           \\
Lumos-1           & 3.6B & 10M & 78.3 & 79.5 & 73.5 & -    & -    & 58.0 & 90.1 & -    & -    & 82.0 & -     \\
Emu3 & 8B         & -    & 81.0 & 84.1 & 68.4 & 79.3 & 59.6 & 62.6 & 86.2 & 44.6 & 68.7 & 88.3 & 37.1        \\
\midrule
\OursRegular      & 1.7B & 24M & 82.4 & 83.4 & 78.5 & 81.4 & 63.1 & 62.2 & 93.4 & 70.6 & 62.1 & 90.7 & 52.3  \\ 
\bottomrule
\end{tabular}}
\end{center}
\end{table}

\textbf{\OursRegular emerges frame-conditioned video generation by accurately modeling the future motion.}
Prior methods typically adapt text-to-image~\citep{MODEL:CONSISTI2V,MODEL:SEINE,MODEL:DYNAMICRAFTER} or text-to-video models with a clean first frame for image-to-video generation.
In contrast, \OursRegular seamlessly integrates asynchronous frame conditions, enabling zero-shot generalization for this task.
As depicted in Table~\ref{tab:res-i2v}, \OursRegular excels in camera control and subject movement versus specialized frame-conditioned models~\citep{MODEL:COSMOS,MODEL:VIDEOMAR,MODEL:WAN,MODEL:PUSA}. 
Our results demonstrate that diffusion forcing effectively generalizes to image-to-video generation,
pushing the boundaries of autoregressive discrete video generation models without causal attention.

\begin{table}[ht!]
\caption{\textbf{Image-to-video evaluation on VBench++.}
To evaluate temporal consistency, we focus on image-to-video (I2V) metrics of 
visual similarity between each generated frame and reference image.
}
\vspace{-3mm}
\label{tab:res-i2v}
\begin{center}
\resizebox{\linewidth}{!}{
\renewcommand\arraystretch{1.1} 
\begin{tabular}{lccccccccccc}
\toprule
Model & \#params & \#videos & 
\makecell{Total\\Score} & \makecell{Quality\\Score} & \makecell{I2V\\Score} & 
\makecell{Dynamic\\Degree} & \makecell{Aesthetic\\Quality} & \makecell{Imaging\\Quality} &
\makecell{Camera\\Motion} & \makecell{I2V Subject\\Consistency} & \makecell{I2V Background\\Consistency} \\
\midrule
\rowcolor{mygray} \multicolumn{12}{l}{$\blacktriangledown$ \emph{Continuous models}}  \\
ConsistI2V        & 2B   & 10M  & 84.1 & 76.2 & 91.9 & 18.6 & 59.0 & 66.9 & 33.9 & 95.8 & 96.0 \\
SEINE             & 3B   & 25M  & 85.5 & 78.4 & 92.7 & 27.1 & 64.6 & 71.4 & 21.0 & 97.2 & 97.0 \\
DynamiCrafter     & 2B   & 10M  & 86.9 & 80.5 & 93.5 & 69.7 & 60.9 & 68.6 & 31.2 & 97.2 & 97.4 \\
\midrule
Cosmos            & 13B  & 100M & 84.2 & 75.8 & 92.6 & 18.7 & 55.8 & 59.9 & 25.4 & 96.0 & 97.4 \\
VideoMAR          & 1.4B & 0.5M & 84.8 & 75.6 & 94.0 & 11.0 & 55.8 & 62.3 & 21.6 & 97.9 & 98.4 \\
\midrule
CogVideoX         & 5B   & -    & 86.7 & 78.6 & 94.8 & 33.2 & 61.9 & 70.0 & 67.7 & 97.2 & 96.7 \\
HunyuanVideo      & 13B  & -    & 86.8 & 78.5 & 95.1 & 22.2 & 62.6 & 70.1 & 49.9 & 98.5 & 97.4 \\
Wan2.1            & 14B  & -    & 86.9 & 80.8 & 92.9 & 51.4 & 64.8 & 70.4 & 34.8 & 97.0 & 96.4 \\
Pusa              & 14B  & -    & 87.3 & 79.8 & 94.8 & 52.6 & 63.2 & 68.3 & 29.5 & 97.6 & 99.2 \\
Step-Video        & 30B  & -    & 88.4 & 81.2 & 95.5 & 48.8 & 62.3 & 70.4 & 49.2 & 97.9 & 98.5 \\
MAGI-1            & 24B  & -    & 89.3 & 82.4 & 96.1 & 68.2 & 64.7 & 69.7 & 50.9 & 98.4 & 99.0 \\
\rowcolor{mygray} \multicolumn{12}{l}{$\blacktriangledown$ \emph{Discrete models}}  \\
Lumos-1           & 3.6B & 10M & 84.7 & 76.1 & 93.3 & -    & -    & 69.2 & -    & 97.4 & 97.4 \\
\midrule
\OursRegular      & 1.7B & 24M & 86.2 & 79.8 & 92.6 & 65.3 & 57.4 & 64.2 & 37.6 & 96.1 & 96.5 \\     
\bottomrule
\end{tabular}}
\end{center}
\end{table}

\textbf{\OursRegular performs on par with the state-of-the-art models in generating high-resolution images.}
We compare \OursRegular against continuous models in Table~\ref{tab:res-t2i}, 
encompassing specialist architectures: SDXL~\citep{MODEL:SDXL}, SD3~\citep{MODEL:SD3}, FLUX~\citep{MODEL:FLUX}, 
SANA~\citep{MODEL:SANA1.5} and NOVA~\citep{MODEL:NOVA}, as well as unified architectures:
Mogao~\citep{MODEL:MOGAO}, Bagel~\citep{MODEL:BAGEL}, OmniGen2~\citep{MODEL:OMNIGEN2} and Show-o2~\citep{MODEL:SHOWO2}.
Through joint modeling of discrete text and visual tokens, \OursRegular demonstrates strong text-image alignment. For example, on the DPG-Bench, \OursRegular reaches a leading overall score with dense text prompts.
This strong performance is consistently sustained on the GenEval when using the rewritten prompts.
At high resolutions, \OursRegular surpasses the autoregressive~\citep{VLM:EMU3,MODEL:INFINITY,MODEL:JANUSPRO} and masked diffusion~\citep{MODEL:MEISSONIC,MODEL:LUMOS} approaches in efficiency, effectively reducing inference steps through iterative refinement while preserving fine-grained detail.

\begin{table}[ht!]
\caption{
\textbf{Text-to-image evaluation on DPG-Bench and GenEval.}
We prefer the DPG-Bench metrics to mitigate potential 
prompt template leakage concerns~\citep{MODEL:RECA} associated with GenEval.
$\dagger$ refers to the methods using rewritten GenEval prompts for clearer position and attribute guidance.
}
\vspace{-3mm}
\label{tab:res-t2i}
\begin{center}
\resizebox{\linewidth}{!}{
\renewcommand{\arraystretch}{1.1}
\begin{tabular}{lccccccccccccc}
\toprule
\multirow{3}{*}{Model} & 
\multicolumn{2}{c}{\textbf{ModelSpec}} & 
\multicolumn{4}{c}{\textbf{DPG-Bench}} &
\multicolumn{7}{c}{\textbf{GenEval}} \\
\cmidrule(lr){2-3}\cmidrule(lr){4-7}\cmidrule(lr){8-14} & 
\#params & \#images & 
Overall & Entity & Attribute & Relation & 
Overall & Single & Two & Counting & Colors & Position & ColorAttr \\
\midrule
\rowcolor{mygray} \multicolumn{14}{l}{$\blacktriangledown$ \emph{Continuous models}}  \\
SDXL                           & 2.6B & -    & 74.7 & 82.4 & 80.9 & 86.8 & 0.55 & 0.98 & 0.44 & 0.39 & 0.85 & 0.15 & 0.23 \\
SD3                            & 2B   & -    & 84.1 & 91.0 & 88.8 & 80.7 & 0.62 & 0.98 & 0.74 & 0.63 & 0.67 & 0.34 & 0.36 \\
FLUX.1-dev                     & 12B  & -    & 84.9 & -    & -    & -    & 0.68 & 0.99 & 0.85 & 0.74 & 0.79 & 0.21 & 0.48 \\
NOVA                           & 1.4B & 600M & 83.0 & 88.7 & 86.4 & 91.9 & 0.71 & 0.99 & 0.91 & 0.62 & 0.85 & 0.33 & 0.56 \\
SANA-1.5\textsuperscript{\dag} & 4.8B & 50M  & 84.7 & -    & -    & -    & 0.81 & 0.99 & 0.93 & 0.86 & 0.84 & 0.59 & 0.65 \\
\midrule
OmniGen2                       & 7B   & -    & 83.6 & 88.8 & 90.2 & 89.4 & 0.80 & 1.00 & 0.95 & 0.64 & 0.88 & 0.55 & 0.76 \\
Mogao\textsuperscript{\dag}    & 7B   & -    & 84.3 & 90.0 & 88.3 & 93.2 & 0.89 & 1.00 & 0.97 & 0.83 & 0.93 & 0.84 & 0.80 \\ 
Bagel                          & 14B  & -    & 85.1 & 90.4 & 91.3 & 90.8 & 0.82 & 0.99 & 0.94 & 0.81 & 0.88 & 0.64 & 0.63 \\
Show-o2\textsuperscript{\dag}  & 7B & 66M    & 86.1 & 91.8 & 90.0 & 91.8 & 0.76 & 1.00 & 0.87 & 0.58 & 0.92 & 0.52 & 0.62 \\
\midrule
\rowcolor{mygray} \multicolumn{14}{l}{$\blacktriangledown$ \emph{Discrete models}}  \\
Show-o                         & 1.3B & 2B  & 67.3 & 75.4 & 78.0 & 84.5 & 0.68 & 0.98 & 0.80 & 0.66 & 0.84 & 0.31 & 0.50 \\
Emu3\textsuperscript{\dag}     & 8B   & -   & 81.6 & 87.2 & 86.3 & 90.6 & 0.66 & 0.99 & 0.81 & 0.42 & 0.80 & 0.49 & 0.45 \\
FUDOKI                         & 1.5B & 13M & 83.6 & 89.7 & 88.1 & 93.7 & 0.77 & 0.96 & 0.85 & 0.56 & 0.88 & 0.68 & 0.67 \\
Janus-Pro                      & 7B   & 72M & 84.2 & 88.9 & 89.4 & 89.3 & 0.80 & 0.99 & 0.89 & 0.59 & 0.90 & 0.79 & 0.66 \\
\midrule
Meissonic                      & 1B & 210M  & -    & -    & -    & -    & 0.54 & 0.99 & 0.66 & 0.42 & 0.86 & 0.10 & 0.22 \\
Lumos-1\textsuperscript{\dag}  & 3.6B & 60M & -    & -    & -    & -    & 0.66 & 0.95 & 0.80 & 0.46 & 0.81 & 0.48 & 0.48 \\
Infinity\textsuperscript{\dag} & 2B   & -   & 83.5 & -    & -    & 90.8 & 0.73 & 0.99 & 0.85 & 0.64 & 0.84 & 0.49 & 0.57 \\
\midrule
\OursRegular \;\;\,(512$\times$320)                    & 1.7B & 30M & 82.5 & 88.3 & 86.4 & 92.9 & 0.64 & 0.99 & 0.83 & 0.47 & 0.83 & 0.30 & 0.41 \\
\OursRegular \;(1024$\times$1024)                      & 1.7B & 30M & 86.0 & 91.5 & 89.6 & 94.7 & 0.68 & 0.99 & 0.92 & 0.63 & 0.86 & 0.25 & 0.40 \\
\OursRegular\!\textsuperscript{\dag}\,(1024$\times$1024) & 1.7B & 30M & -    & -    & -    & -    & 0.80 & 1.00 & 0.92 & 0.64 & 0.89 & 0.67 & 0.69 \\
\bottomrule
\end{tabular}}
\end{center}
\vspace{-5mm}
\end{table}

\subsection{Ablation Study}\label{sec:ablation}

\begin{figure}[ht!]
\resizebox{\textwidth}{!}{
\setlength{\tabcolsep}{0.5mm}
\begin{tabular}{ccc}
\multicolumn{3}{c}{\qquad\quad\includegraphics[width=1.45\linewidth]{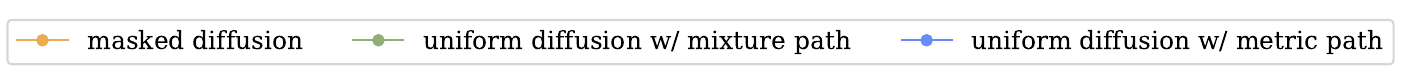}} \\
\includegraphics[width=0.55\linewidth]{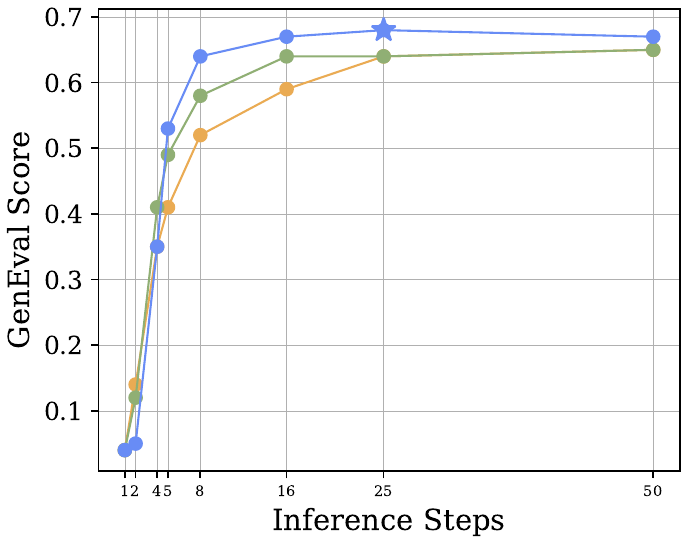} &
\includegraphics[width=0.55\linewidth]{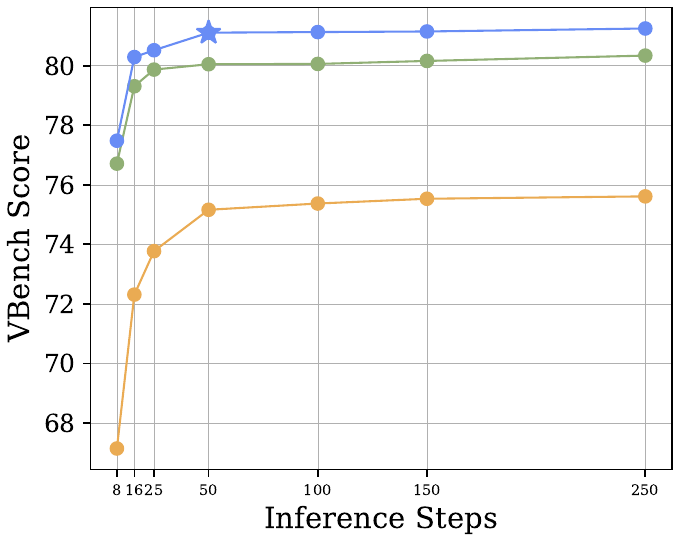} &
\includegraphics[width=0.55\linewidth]{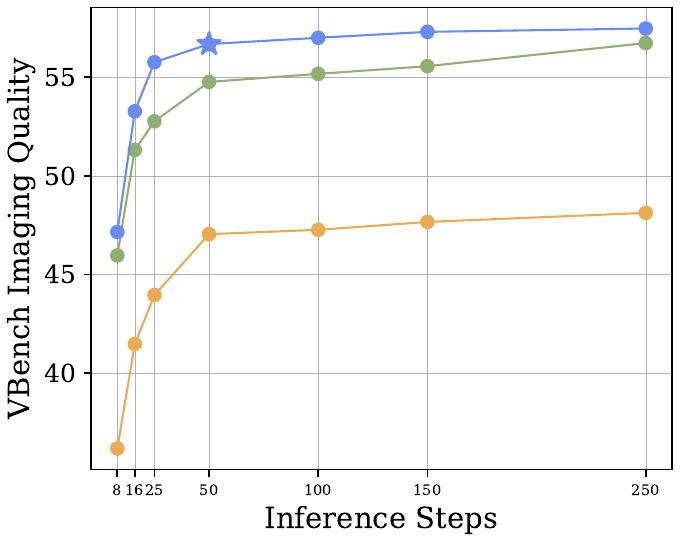} \\
\multicolumn{1}{r}{\large{(a) Text-to-image performance on GenEval.}} & 
\multicolumn{2}{c}{\large{(b) Text-to-video performance on VBench.}} \\
\end{tabular}}
\caption{\textbf{Sampling performance across inference steps.}
Using the Cosmos tokenizer, we evaluate the image samples at 256$\times$256 ($\sim$1K tokens)
and the video samples at 25$\times$384$\times$240 ($\sim$10K tokens).
}
\label{fig:ablation_refinement}
\vspace{-1em}
\end{figure}

\textbf{Effectiveness of iterative refinement for visual generation.}
Discrete diffusion models inherently incur elevated sampling errors, as exhibited in prior studies~\citep{ALGO:VQERROR,ALGO:MDMERROR}.
To systematically investigate this issue in image and video generation, we train three variants of the discrete diffusion model, assessing performance across insufficient and excessive sampling regimes.
Figure~\ref{fig:ablation_refinement} compares key performance metrics of text-to-image models on GenEval and text-to-video models on VBench, with all models evaluated after being trained for an identical number of iterations.
In the image generation task, which is characterized by low structural redundancy, all three models can generate feasible images within the conventional 25 inference steps.
Without iterative refinement, reducing the number of steps substantially decreases the GenEval score in masked diffusion sampling.
As we progress into video generation, a task rich in contextual redundancy, it becomes essential to correct sampling errors at each step, ensuring temporal coherence and visual fidelity across frames.

\begin{figure}[ht!]
\vspace{-1em}
\resizebox{\textwidth}{!}{
\setlength{\tabcolsep}{0.5mm}
\begin{tabular}{cc}
\multicolumn{2}{c}{\qquad\quad\includegraphics[width=1.02\linewidth]{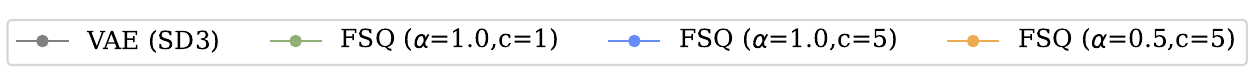}} \\
\includegraphics[width=0.55\linewidth]{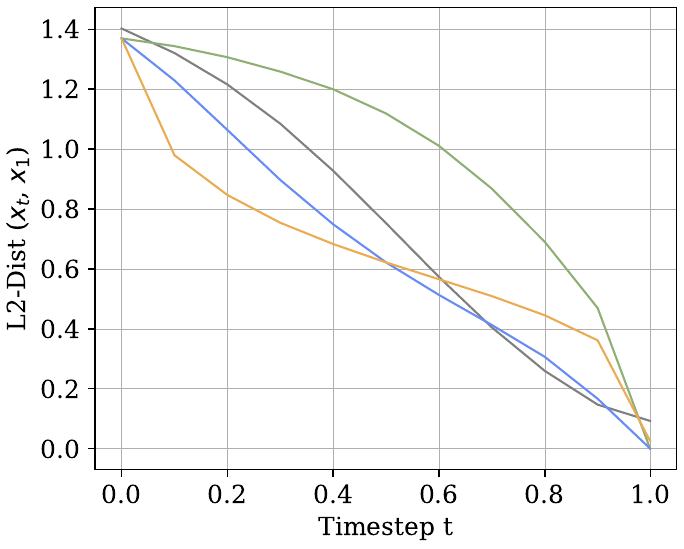} &
\includegraphics[width=0.55\linewidth]{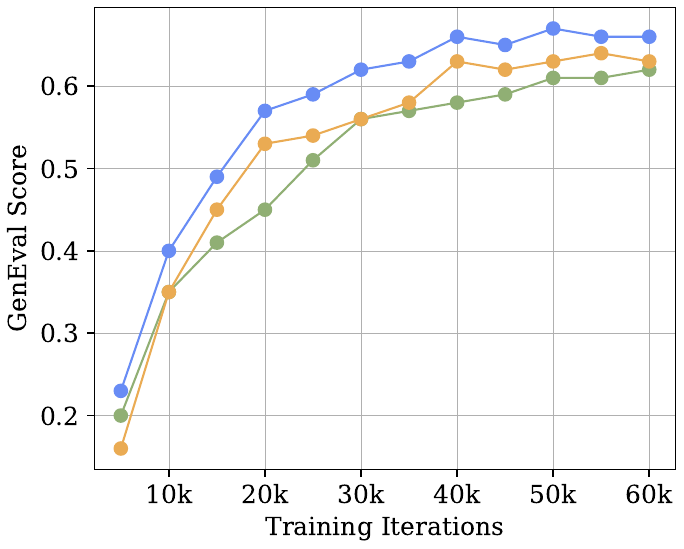} 
\end{tabular}}
\caption{\textbf{Sampling performance of different paths.} We evaluate the image samples at 256$\times$256.}
\vspace{-1em}
\label{fig:ablation_path}
\end{figure}

\textbf{Effectiveness of path linearity for uniform diffusion.}
As shown in Figure~\ref{fig:ablation_path}, the left plot shows the average Euclidean distance between the embedding of noisy images and the clean image using 10K images sampled from the training set.
We compute the Pearson correlation coefficients between the Euclidean distance and the timestep, which are \textcolor{gray}{-0.995}, \textcolor[HTML]{90AF74}{-0.921}, \textcolor[HTML]{688CF5}{-0.997}, and \textcolor[HTML]{EAAB53}{-0.949}. 
We find that the choice of the probability path is significantly influenced by the values of $c$ and $\alpha$, which has a substantial impact on the model performance.
To determine the optimal values for $c$ and $\alpha$, we draw inspiration from the continuous diffusion model \textcolor{gray}{SD3}~\citep{MODEL:SD3}, where the relationship between $t$ and $d(x_t, x_1)$ demonstrates a strong linear correlation. 
This insight guides our approach to calibrating $c$ and $\alpha$ to effectively reach the limits of model performance for different vision tokenizers.

\begin{figure}[ht!]
\resizebox{\textwidth}{!}{
\setlength{\tabcolsep}{0.5mm}
\begin{tabular}{cc}
\includegraphics[width=0.55\linewidth]{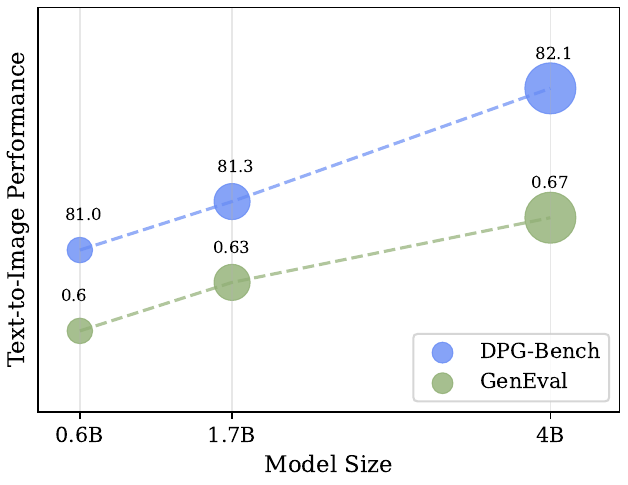} &
\includegraphics[width=0.55\linewidth]{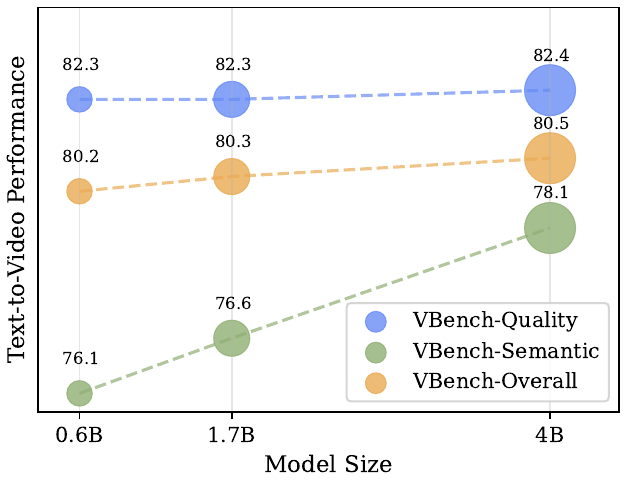} \\
\end{tabular}}
\caption{\textbf{Sampling performance of different model sizes.}
All models are trained for the same epoch count as in the main experiments and evaluated on 256$\times$256 images and 25$\times$384$\times$240 videos.
}
\vspace{-0.5em}
\label{fig:ablation_scaling}
\end{figure}

\textbf{Effectiveness of model size for uniform diffusion.}
To study the scaling properties of \OursRegular models, we train three variants that are initialized from Qwen3 models with 0.6B, 1.7B, and 4B parameters.
Figure~\ref{fig:ablation_scaling} compares the performance of different model sizes on DPG-Bench, GenEval, and VBench, with all models trained for the same epoch count as in Sec.~\ref{sec:experiment:main}.
We find that increasing model size considerably enhances semantic performance across both text-to-image and text-to-video evaluations but does not significantly improve generation quality.
This suggests that while larger models better capture high-level semantics and align more accurately with text prompts, the fidelity of the generated outputs may ultimately be constrained by the representation capacity of the discrete vision tokenizer. 

\begin{figure}[ht!]
\resizebox{\textwidth}{!}{
\setlength{\tabcolsep}{0.8mm}
\begin{tabular}{ccc}
\multicolumn{3}{c}{\qquad\quad\includegraphics[width=1.45\linewidth]{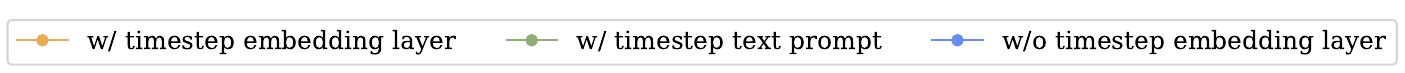}} \\
\includegraphics[width=0.55\linewidth]{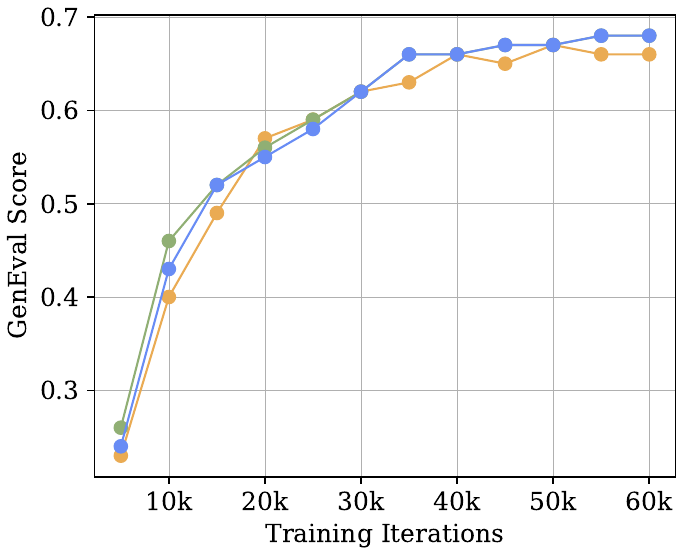} &
\includegraphics[width=0.55\linewidth]{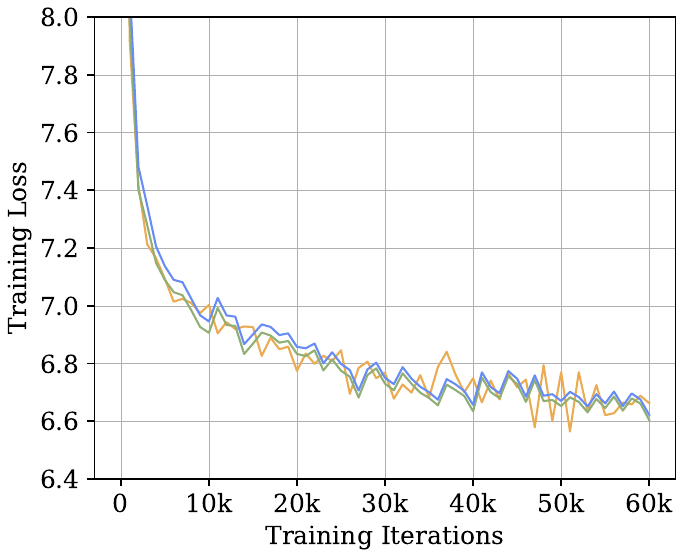} &
\includegraphics[width=0.55\linewidth]{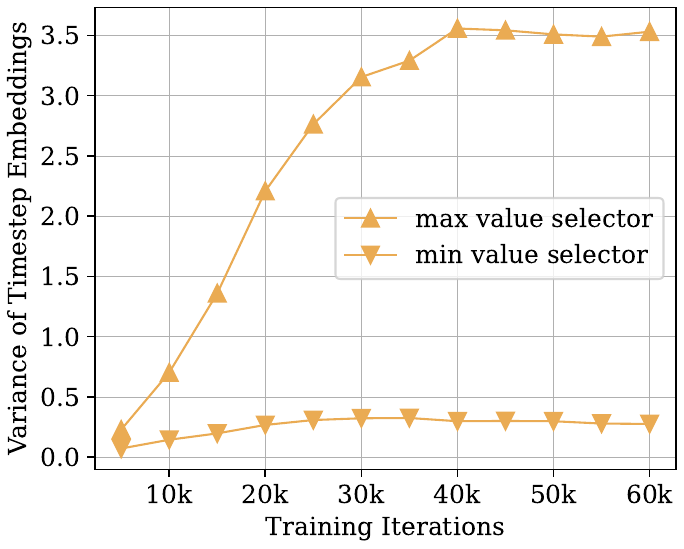} \\
\end{tabular}}
\caption{\textbf{Model metrics across training iterations.}
We sample 256$\times$256 images for evaluation.
}
\vspace{-1.5em}
\label{fig:ablation_timestep}
\end{figure}

\textbf{Effectiveness of timestep conditioning for uniform diffusion.}
Recent work explores time-agnostic (i.e., noise-unconditional) methods for both continuous diffusion~\citep{ALGO:NOISECOND-CFM,MODEL:FUSEDIT} and masked diffusion~\citep{ALGO:NOISECOND-MDM,MODEL:RADD}, effectively narrowing the architectural gap between diffusion transformers (DiTs) and LLMs.
In this context, we analyze whether timestep conditioning remains indispensable for uniform diffusion. The results are illustrated in Figure~\ref{fig:ablation_timestep}.
Specifically, we train three model variants with distinct conditioning strategies and evaluate GenEval across training iterations.
After one epoch ($\sim$30K iterations), embedding or prompting with the timestep provides no measurable benefit.
Notably, timestep embedding could degrade performance as their variance increases, potentially disrupting token embedding and compromising training stability.

\begin{figure}[ht!]
\resizebox{\textwidth}{!}{
\setlength{\tabcolsep}{0.1mm}
\begin{tabular}{cc}
\includegraphics[width=0.562\linewidth]{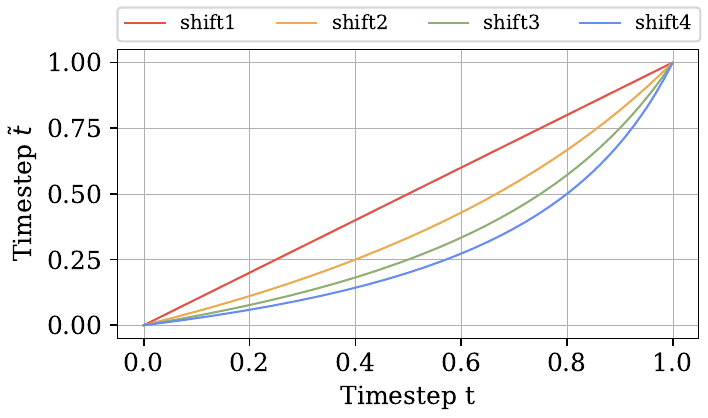} &
\includegraphics[width=0.5\linewidth]{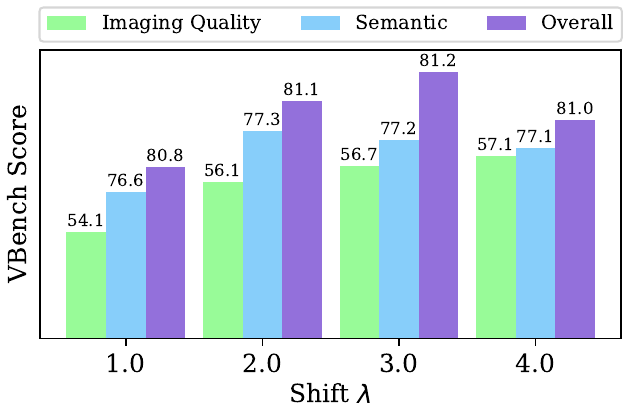} \\
\multicolumn{2}{c}{\includegraphics[width=1.12\linewidth]{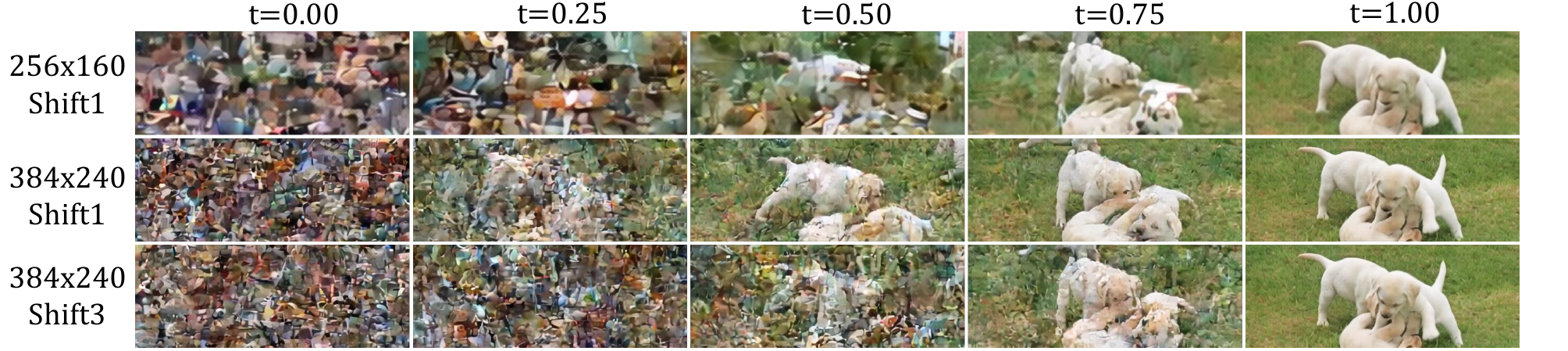}} \\
\end{tabular}}
\caption{\textbf{Timestep shifting across SNR schedules.}
We sample 25$\times$384$\times$240 videos for evaluation.
}
\label{fig:ablation_shift}
\end{figure}

\textbf{Effectiveness of timestep shifting for video generation.}
As outlined in Section~\ref{sec:experiment:setup}, our probability path is designed to maximize the time interval. In line with continuous models~\citep{MODEL:HUNYUANVIDEO,MODEL:WAN,MODEL:LUMINAVIDEO}, the optimal SNR schedule should be tailored with video size.
To study the impact of the SNR schedule on video generation, we train four text-to-video models with divergent timestep shifting and evaluate their performance using the respective value on VBench.
Figure~\ref{fig:ablation_shift} presents our shifting schedules, accompanied by their evaluation metrics and visualizations.
Surprisingly, the shifting strategy proposed by~\citet{MODEL:SD3} demonstrates strong effectiveness for uniform diffusion, empowering \OursRegular to match the performance of its continuous counterparts.

\section{Conclusion}
In this work, we revisited discrete generative modeling for video synthesis and introduced \Ours, 
a uniform diffusion framework with a metric path that bridges discrete and continuous paradigms.
\OursRegular employs two key innovations, a linearized metric path and a resolution-dependent timestep shifting strategy, to provide fine-grained control over perturbations for long sequences. 
On top of this, our asynchronous temporal scheduling strategy enables multi-task video generation in one model.
Extensive experiments show that \OursRegular not only consistently outperforms existing discrete methods but also achieves highly competitive results compared to state-of-the-art continuous diffusion models.
We contend that this work represents a significant step toward unifying discrete and continuous paradigms and provides a promising direction for scalable, versatile, and efficient video generation.

\textbf{Acknowledgements.}
This work was supported in part by the National Natural Science Foundation of China (No. 62320106010, No. U21B2042). We thank Jin Wang and Jiahao Wang (HKU) for their help with algorithm implementation and early feedback. We are grateful to Zhengxiong Luo for insightful comments on the algorithm design during the initial phase. We also thank Yuanzhi Zhu (École Polytechnique), Shilin Lu (NTU), Honghao Chen (CASIA) and Jinming Wu for their helpful suggestions and editorial improvements on the manuscript. We appreciate the stimulating discussions with Chengyuan Wang (BAAI-Vision) and acknowledge the support from other colleagues at BAAI.

\clearpage
\bibliography{reference}

\begin{thebibliography}{97}
\providecommand{\natexlab}[1]{#1}
\providecommand{\url}[1]{\texttt{#1}}
\expandafter\ifx\csname urlstyle\endcsname\relax
  \providecommand{\doi}[1]{doi: #1}\else
  \providecommand{\doi}{doi: \begingroup \urlstyle{rm}\Url}\fi

\bibitem[Agarwal et~al.(2025)Agarwal, Ali, Bala, Balaji, Barker, Cai,
  Chattopadhyay, Chen, Cui, Ding, et~al.]{MODEL:COSMOS}
Niket Agarwal, Arslan Ali, Maciej Bala, Yogesh Balaji, Erik Barker, Tiffany
  Cai, Prithvijit Chattopadhyay, Yongxin Chen, Yin Cui, Yifan Ding, et~al.
\newblock Cosmos world foundation model platform for physical ai.
\newblock \emph{arXiv preprint arXiv:2501.03575}, 2025.

\bibitem[Bai et~al.(2025)Bai, Ye, Chow, Song, Li, Dong, Zhu, and
  Yan]{MODEL:MEISSONIC}
Jinbin Bai, Tian Ye, Wei Chow, Enxin Song, Xiangtai Li, Zhen Dong, Lei Zhu, and
  Shuicheng Yan.
\newblock Meissonic: Revitalizing masked generative transformers for efficient
  high-resolution text-to-image synthesis.
\newblock In \emph{ICLR}, 2025.

\bibitem[Baldridge et~al.(2024)Baldridge, Bauer, Bhutani, Brichtova, Bunner,
  Chan, Chen, Dieleman, Du, Eaton-Rosen, et~al.]{MODEL:IMAGEN3}
Jason Baldridge, Jakob Bauer, Mukul Bhutani, Nicole Brichtova, Andrew Bunner,
  Kelvin Chan, Yichang Chen, Sander Dieleman, Yuqing Du, Zach Eaton-Rosen,
  et~al.
\newblock Imagen 3.
\newblock \emph{arXiv preprint arXiv:2408.07009}, 2024.

\bibitem[Batifol et~al.(2025)Batifol, Blattmann, Boesel, Consul, Diagne,
  Dockhorn, English, English, Esser, et~al.]{MODEL:FLUX}
Stephen Batifol, Andreas Blattmann, Frederic Boesel, Saksham Consul, Cyril
  Diagne, Tim Dockhorn, Jack English, Zion English, Patrick Esser, et~al.
\newblock Flux. 1 kontext: Flow matching for in-context image generation and
  editing in latent space.
\newblock \emph{arXiv preprint arXiv:2506.15742}, 2025.

\bibitem[Betker et~al.(2023)Betker, Goh, Jing, Brooks, Wang, Li, Ouyang,
  Zhuang, Lee, Guo, et~al.]{MODEL:DALLE3}
James Betker, Gabriel Goh, Li~Jing, Tim Brooks, Jianfeng Wang, Linjie Li, Long
  Ouyang, Juntang Zhuang, Joyce Lee, Yufei Guo, et~al.
\newblock Improving image generation with better captions, 2023.

\bibitem[Brock et~al.(2019)Brock, Donahue, and Simonyan]{MODEL:BIGGAN}
Andrew Brock, Jeff Donahue, and Karen Simonyan.
\newblock Large scale gan training for high fidelity natural image synthesis.
\newblock In \emph{ICLR}, 2019.

\bibitem[Brooks et~al.(2024)Brooks, Peebles, Holmes, DePue, Guo, Jing, Schnurr,
  Taylor, Luhman, Luhman, et~al.]{MODEL:SORA}
Tim Brooks, Bill Peebles, Connor Holmes, Will DePue, Yufei Guo, Li~Jing, David
  Schnurr, Joe Taylor, Troy Luhman, Eric Luhman, et~al.
\newblock Video generation models as world simulators, 2024.

\bibitem[Brown et~al.(2020)Brown, Mann, Ryder, Subbiah, Kaplan, Dhariwal,
  Neelakantan, Shyam, Sastry, Askell, et~al.]{LLM:GPT3}
Tom Brown, Benjamin Mann, Nick Ryder, Melanie Subbiah, Jared~D Kaplan, Prafulla
  Dhariwal, Arvind Neelakantan, Pranav Shyam, Girish Sastry, Amanda Askell,
  et~al.
\newblock Language models are few-shot learners.
\newblock In \emph{NeurIPS}, 2020.

\bibitem[Byeon et~al.(2022)Byeon, Park, Kim, Lee, Baek, and Kim]{DATASET:COYO}
Minwoo Byeon, Beomhee Park, Haecheon Kim, Sungjun Lee, Woonhyuk Baek, and
  Saehoon Kim.
\newblock Coyo-700m: Image-text pair dataset, 2022.

\bibitem[Chang et~al.(2022)Chang, Zhang, Jiang, Liu, and
  Freeman]{MODEL:MASKGIT}
Huiwen Chang, Han Zhang, Lu~Jiang, Ce~Liu, and William~T Freeman.
\newblock Maskgit: Masked generative image transformer.
\newblock In \emph{CVPR}, 2022.

\bibitem[Chang et~al.(2023)Chang, Zhang, Barber, Maschinot, Lezama, Jiang,
  Yang, Murphy, Freeman, Rubinstein, et~al.]{MODEL:MUSE}
Huiwen Chang, Han Zhang, Jarred Barber, AJ~Maschinot, Jose Lezama, Lu~Jiang,
  Ming-Hsuan Yang, Kevin Murphy, William~T Freeman, Michael Rubinstein, et~al.
\newblock Muse: Text-to-image generation via masked generative transformers.
\newblock In \emph{ICML}, 2023.

\bibitem[Chen et~al.(2024{\natexlab{a}})Chen, Mart{\'\i}~Mons{\'o}, Du,
  Simchowitz, Tedrake, and Sitzmann]{ALGO:DForcing}
Boyuan Chen, Diego Mart{\'\i}~Mons{\'o}, Yilun Du, Max Simchowitz, Russ
  Tedrake, and Vincent Sitzmann.
\newblock Diffusion forcing: Next-token prediction meets full-sequence
  diffusion.
\newblock In \emph{NeurIPS}, 2024{\natexlab{a}}.

\bibitem[Chen et~al.(2025{\natexlab{a}})Chen, Lin, Yang, Lin, Zhu, Fan, Zhang,
  Chen, Chen, Ma, et~al.]{MODEL:SKYREELSV2}
Guibin Chen, Dixuan Lin, Jiangping Yang, Chunze Lin, Junchen Zhu, Mingyuan Fan,
  Hao Zhang, Sheng Chen, Zheng Chen, Chengcheng Ma, et~al.
\newblock Skyreels-v2: Infinite-length film generative model.
\newblock \emph{arXiv preprint arXiv:2504.13074}, 2025{\natexlab{a}}.

\bibitem[Chen et~al.(2025{\natexlab{b}})Chen, Wu, Liu, Pan, Liu, Xie, Yu, and
  Ruan]{MODEL:JANUSPRO}
Xiaokang Chen, Zhiyu Wu, Xingchao Liu, Zizheng Pan, Wen Liu, Zhenda Xie,
  Xingkai Yu, and Chong Ruan.
\newblock Janus-pro: Unified multimodal understanding and generation with data
  and model scaling.
\newblock \emph{arXiv preprint arXiv:2501.17811}, 2025{\natexlab{b}}.

\bibitem[Chen et~al.(2024{\natexlab{b}})Chen, Wang, Zhang, Zhuang, Ma, Yu,
  Wang, Lin, Qiao, and Liu]{MODEL:SEINE}
Xinyuan Chen, Yaohui Wang, Lingjun Zhang, Shaobin Zhuang, Xin Ma, Jiashuo Yu,
  Yali Wang, Dahua Lin, Yu~Qiao, and Ziwei Liu.
\newblock Seine: Short-to-long video diffusion model for generative transition
  and prediction.
\newblock In \emph{ICLR}, 2024{\natexlab{b}}.

\bibitem[Dehghani et~al.(2023)Dehghani, Djolonga, Mustafa, Padlewski, Heek,
  Gilmer, Steiner, Caron, Geirhos, Alabdulmohsin, et~al.]{MODEL:VIT22B}
Mostafa Dehghani, Josip Djolonga, Basil Mustafa, Piotr Padlewski, Jonathan
  Heek, Justin Gilmer, Andreas~Peter Steiner, Mathilde Caron, Robert Geirhos,
  Ibrahim Alabdulmohsin, et~al.
\newblock Scaling vision transformers to 22 billion parameters.
\newblock In \emph{ICML}, 2023.

\bibitem[Deng et~al.(2025{\natexlab{a}})Deng, Zhu, Li, Gou, Li, Wang, Zhong,
  Yu, Nie, Song, et~al.]{MODEL:BAGEL}
Chaorui Deng, Deyao Zhu, Kunchang Li, Chenhui Gou, Feng Li, Zeyu Wang, Shu
  Zhong, Weihao Yu, Xiaonan Nie, Ziang Song, et~al.
\newblock Emerging properties in unified multimodal pretraining.
\newblock \emph{arXiv preprint arXiv:2505.14683}, 2025{\natexlab{a}}.

\bibitem[Deng et~al.(2025{\natexlab{b}})Deng, Pan, Diao, Luo, Cui, Lu, Shan,
  Qi, and Wang]{MODEL:NOVA}
Haoge Deng, Ting Pan, Haiwen Diao, Zhengxiong Luo, Yufeng Cui, Huchuan Lu,
  Shiguang Shan, Yonggang Qi, and Xinlong Wang.
\newblock Autoregressive video generation without vector quantization.
\newblock In \emph{ICLR}, 2025{\natexlab{b}}.

\bibitem[Diao et~al.(2024)Diao, Cui, Li, Wang, Lu, and Wang]{VLM:EVE}
Haiwen Diao, Yufeng Cui, Xiaotong Li, Yueze Wang, Huchuan Lu, and Xinlong Wang.
\newblock Unveiling encoder-free vision-language models.
\newblock In \emph{NeurIPS}, 2024.

\bibitem[Ding et~al.(2021)Ding, Yang, Hong, Zheng, Zhou, Yin, Lin, Zou, Shao,
  Yang, et~al.]{MODEL:COGVIEW1}
Ming Ding, Zhuoyi Yang, Wenyi Hong, Wendi Zheng, Chang Zhou, Da~Yin, Junyang
  Lin, Xu~Zou, Zhou Shao, Hongxia Yang, et~al.
\newblock Cogview: Mastering text-to-image generation via transformers.
\newblock In \emph{NeurIPS}, 2021.

\bibitem[Dinh et~al.(2014)Dinh, Krueger, and Bengio]{MODEL:NICE}
Laurent Dinh, David Krueger, and Yoshua Bengio.
\newblock Nice: Non-linear independent components estimation.
\newblock \emph{arXiv preprint arXiv:1410.8516}, 2014.

\bibitem[Dinh et~al.(2017)Dinh, Sohl-Dickstein, and Bengio]{MODEL:REALNVP}
Laurent Dinh, Jascha Sohl-Dickstein, and Samy Bengio.
\newblock Density estimation using real nvp.
\newblock In \emph{ICLR}, 2017.

\bibitem[Esser et~al.(2021)Esser, Rombach, and Ommer]{TOKENIZER:VQGAN}
Patrick Esser, Robin Rombach, and Bjorn Ommer.
\newblock Taming transformers for high-resolution image synthesis.
\newblock In \emph{CVPR}, 2021.

\bibitem[Esser et~al.(2024)Esser, Kulal, Blattmann, Entezari, M{\"u}ller,
  Saini, Levi, Lorenz, Sauer, Boesel, et~al.]{MODEL:SD3}
Patrick Esser, Sumith Kulal, Andreas Blattmann, Rahim Entezari, Jonas
  M{\"u}ller, Harry Saini, Yam Levi, Dominik Lorenz, Axel Sauer, Frederic
  Boesel, et~al.
\newblock Scaling rectified flow transformers for high-resolution image
  synthesis.
\newblock In \emph{ICML}, 2024.

\bibitem[Fan et~al.(2025)Fan, Si, Song, Yang, He, Zhuo, Huang, Dong, He, Pan,
  et~al.]{MODEL:VCHITECT2}
Weichen Fan, Chenyang Si, Junhao Song, Zhenyu Yang, Yinan He, Long Zhuo, Ziqi
  Huang, Ziyue Dong, Jingwen He, Dongwei Pan, et~al.
\newblock Vchitect-2.0: Parallel transformer for scaling up video diffusion
  models.
\newblock \emph{arXiv preprint arXiv:2501.08453}, 2025.

\bibitem[Feng et~al.(2025)Feng, Geng, Guan, Wu, Wang, and He]{ALGO:MDMERROR}
Guhao Feng, Yihan Geng, Jian Guan, Wei Wu, Liwei Wang, and Di~He.
\newblock Theoretical benefit and limitation of diffusion language model.
\newblock In \emph{NeurIPS}, 2025.

\bibitem[Gadre et~al.(2024)Gadre, Ilharco, Fang, Hayase, Smyrnis, Nguyen,
  Marten, Wortsman, Ghosh, Zhang, et~al.]{DATASET:DATACOMP}
Samir~Yitzhak Gadre, Gabriel Ilharco, Alex Fang, Jonathan Hayase, Georgios
  Smyrnis, Thao Nguyen, Ryan Marten, Mitchell Wortsman, Dhruba Ghosh, Jieyu
  Zhang, et~al.
\newblock Datacomp: In search of the next generation of multimodal datasets.
\newblock In \emph{NeurIPS}, 2024.

\bibitem[Gao et~al.(2025{\natexlab{a}})Gao, Gong, Guo, Hou, Lai, Li, Li, Lian,
  Liao, Liu, et~al.]{MODEL:SEEDREAM}
Yu~Gao, Lixue Gong, Qiushan Guo, Xiaoxia Hou, Zhichao Lai, Fanshi Li, Liang Li,
  Xiaochen Lian, Chao Liao, Liyang Liu, et~al.
\newblock Seedream 3.0 technical report.
\newblock \emph{arXiv preprint arXiv:2504.11346}, 2025{\natexlab{a}}.

\bibitem[Gao et~al.(2025{\natexlab{b}})Gao, Guo, Hoang, Huang, Jiang, Kong, Li,
  Li, Li, Li, et~al.]{MODEL:SEEDANCE}
Yu~Gao, Haoyuan Guo, Tuyen Hoang, Weilin Huang, Lu~Jiang, Fangyuan Kong, Huixia
  Li, Jiashi Li, Liang Li, Xiaojie Li, et~al.
\newblock Seedance 1.0: Exploring the boundaries of video generation models.
\newblock \emph{arXiv preprint arXiv:2506.09113}, 2025{\natexlab{b}}.

\bibitem[Gat et~al.(2024)Gat, Remez, Shaul, Kreuk, Chen, Synnaeve, Adi, and
  Lipman]{ALGO:DFM}
Itai Gat, Tal Remez, Neta Shaul, Felix Kreuk, Ricky~TQ Chen, Gabriel Synnaeve,
  Yossi Adi, and Yaron Lipman.
\newblock Discrete flow matching.
\newblock In \emph{NeurIPS}, 2024.

\bibitem[Ghosh et~al.(2024)Ghosh, Hajishirzi, and Schmidt]{EVAL:GENEVAL}
Dhruba Ghosh, Hannaneh Hajishirzi, and Ludwig Schmidt.
\newblock Geneval: An object-focused framework for evaluating text-to-image
  alignment.
\newblock In \emph{NeurIPS}, 2024.

\bibitem[Goodfellow et~al.(2014)Goodfellow, Pouget-Abadie, Mirza, Xu,
  Warde-Farley, Ozair, Courville, and Bengio]{ALGO:GAN}
Ian~J Goodfellow, Jean Pouget-Abadie, Mehdi Mirza, Bing Xu, David Warde-Farley,
  Sherjil Ozair, Aaron Courville, and Yoshua Bengio.
\newblock Generative adversarial nets.
\newblock In \emph{NeurIPS}, 2014.

\bibitem[Han et~al.(2025)Han, Liu, Jiang, Yan, Zhang, Yuan, Peng, and
  Liu]{MODEL:INFINITY}
Jian Han, Jinlai Liu, Yi~Jiang, Bin Yan, Yuqi Zhang, Zehuan Yuan, Bingyue Peng,
  and Xiaobing Liu.
\newblock Infinity: Scaling bitwise autoregressive modeling for high-resolution
  image synthesis.
\newblock In \emph{CVPR}, 2025.

\bibitem[Ho \& Salimans(2022)Ho and Salimans]{ALGO:CFG}
Jonathan Ho and Tim Salimans.
\newblock Classifier-free diffusion guidance.
\newblock \emph{arXiv preprint arXiv:2207.12598}, 2022.

\bibitem[Ho et~al.(2020)Ho, Jain, and Abbeel]{ALGO:DDPM}
Jonathan Ho, Ajay Jain, and Pieter Abbeel.
\newblock Denoising diffusion probabilistic models.
\newblock In \emph{NeurIPS}, 2020.

\bibitem[Hu et~al.(2024)Hu, Wang, Fang, Fu, Cheng, and Yu]{EVAL:DPGBENCH}
Xiwei Hu, Rui Wang, Yixiao Fang, Bin Fu, Pei Cheng, and Gang Yu.
\newblock Ella: Equip diffusion models with llm for enhanced semantic
  alignment.
\newblock \emph{arXiv preprint arXiv:2403.05135}, 2024.

\bibitem[Huang et~al.(2024{\natexlab{a}})Huang, He, Yu, Zhang, Si, Jiang,
  Zhang, Wu, Jin, Chanpaisit, et~al.]{EVAL:VBENCH}
Ziqi Huang, Yinan He, Jiashuo Yu, Fan Zhang, Chenyang Si, Yuming Jiang, Yuanhan
  Zhang, Tianxing Wu, Qingyang Jin, Nattapol Chanpaisit, et~al.
\newblock Vbench: Comprehensive benchmark suite for video generative models.
\newblock In \emph{CVPR}, 2024{\natexlab{a}}.

\bibitem[Huang et~al.(2024{\natexlab{b}})Huang, Zhang, Xu, He, Yu, Dong, Ma,
  Chanpaisit, Si, Jiang, et~al.]{EVAL:VBENCHPLUS}
Ziqi Huang, Fan Zhang, Xiaojie Xu, Yinan He, Jiashuo Yu, Ziyue Dong, Qianli Ma,
  Nattapol Chanpaisit, Chenyang Si, Yuming Jiang, et~al.
\newblock Vbench++: Comprehensive and versatile benchmark suite for video
  generative models.
\newblock \emph{arXiv preprint arXiv:2411.13503}, 2024{\natexlab{b}}.

\bibitem[Jin et~al.(2025)Jin, Sun, Li, Xu, Xu, Jiang, Zhuang, Huang, Song, MU,
  and Lin]{MODEL:PYRAMIDFLOW}
Yang Jin, Zhicheng Sun, Ningyuan Li, Kun Xu, Kun Xu, Hao Jiang, Nan Zhuang,
  Quzhe Huang, Yang Song, Yadong MU, and Zhouchen Lin.
\newblock Pyramidal flow matching for efficient video generative modeling.
\newblock In \emph{ICLR}, 2025.

\bibitem[Kalchbrenner et~al.(2017)Kalchbrenner, Oord, Simonyan, Danihelka,
  Vinyals, Graves, and Kavukcuoglu]{MODEL:VIDEOPIXEL}
Nal Kalchbrenner, A{\"a}ron Oord, Karen Simonyan, Ivo Danihelka, Oriol Vinyals,
  Alex Graves, and Koray Kavukcuoglu.
\newblock Video pixel networks.
\newblock In \emph{ICML}, 2017.

\bibitem[Karras et~al.(2020)Karras, Laine, Aittala, Hellsten, Lehtinen, and
  Aila]{MODEL:STYLEGAN}
Tero Karras, Samuli Laine, Miika Aittala, Janne Hellsten, Jaakko Lehtinen, and
  Timo Aila.
\newblock Analyzing and improving the image quality of stylegan.
\newblock In \emph{CVPR}, 2020.

\bibitem[Kingma \& Welling(2014)Kingma and Welling]{TOKENIZER:VAE}
Diederik~P Kingma and Max Welling.
\newblock Auto-encoding variational bayes.
\newblock In \emph{ICLR}, 2014.

\bibitem[Kondratyuk et~al.(2024)Kondratyuk, Yu, Gu, Lezama, Huang, Schindler,
  Hornung, Birodkar, Yan, Chiu, et~al.]{MODEL:VIDEOPOET}
Dan Kondratyuk, Lijun Yu, Xiuye Gu, Jose Lezama, Jonathan Huang, Grant
  Schindler, Rachel Hornung, Vighnesh Birodkar, Jimmy Yan, Ming-Chang Chiu,
  et~al.
\newblock Videopoet: A large language model for zero-shot video generation.
\newblock In \emph{ICML}, 2024.

\bibitem[Kong et~al.(2024)Kong, Tian, Zhang, Min, Dai, Zhou, Xiong, Li, Wu,
  Zhang, et~al.]{MODEL:HUNYUANVIDEO}
Weijie Kong, Qi~Tian, Zijian Zhang, Rox Min, Zuozhuo Dai, Jin Zhou, Jiangfeng
  Xiong, Xin Li, Bo~Wu, Jianwei Zhang, et~al.
\newblock Hunyuanvideo: A systematic framework for large video generative
  models.
\newblock \emph{arXiv preprint arXiv:2412.03603}, 2024.

\bibitem[Kuaishou(2024)]{MODEL:KLING}
Kuaishou.
\newblock Kling ai, 2024.
\newblock URL \url{https://klingai.com}.

\bibitem[Li et~al.(2024)Li, Tian, Li, Deng, and He]{MODEL:MAR}
Tianhong Li, Yonglong Tian, He~Li, Mingyang Deng, and Kaiming He.
\newblock Autoregressive image generation without vector quantization.
\newblock In \emph{NeurIPS}, 2024.

\bibitem[Liao et~al.(2025)Liao, Liu, Wang, Luo, Zhang, Zhao, Wu, Li, Tian, and
  Huang]{MODEL:MOGAO}
Chao Liao, Liyang Liu, Xun Wang, Zhengxiong Luo, Xinyu Zhang, Wenliang Zhao,
  Jie Wu, Liang Li, Zhi Tian, and Weilin Huang.
\newblock Mogao: An omni foundation model for interleaved multi-modal
  generation.
\newblock \emph{arXiv preprint arXiv:2505.05472}, 2025.

\bibitem[Lin et~al.(2024)Lin, Ge, Cheng, Li, Zhu, Wang, He, Ye, Yuan, Chen,
  et~al.]{MODEL:OPENSORAPLAN}
Bin Lin, Yunyang Ge, Xinhua Cheng, Zongjian Li, Bin Zhu, Shaodong Wang, Xianyi
  He, Yang Ye, Shenghai Yuan, Liuhan Chen, et~al.
\newblock Open-sora plan: Open-source large video generation model.
\newblock \emph{arXiv preprint arXiv:2412.00131}, 2024.

\bibitem[Liu et~al.(2025{\natexlab{a}})Liu, Li, Liu, Li, Wang, Li, Qin, Liu,
  Xin, Li, et~al.]{MODEL:LUMINAVIDEO}
Dongyang Liu, Shicheng Li, Yutong Liu, Zhen Li, Kai Wang, Xinyue Li, Qi~Qin,
  Yufei Liu, Yi~Xin, Zhongyu Li, et~al.
\newblock Lumina-video: Efficient and flexible video generation with
  multi-scale next-dit.
\newblock \emph{arXiv preprint arXiv:2502.06782}, 2025{\natexlab{a}}.

\bibitem[Liu et~al.(2025{\natexlab{b}})Liu, Ren, Artola, Hu, Cun, Zhao, Zhao,
  Chan, Zhang, Liu, et~al.]{MODEL:PUSA}
Yaofang Liu, Yumeng Ren, Aitor Artola, Yuxuan Hu, Xiaodong Cun, Xiaotong Zhao,
  Alan Zhao, Raymond~H Chan, Suiyun Zhang, Rui Liu, et~al.
\newblock Pusa v1.0: Surpassing wan-i2v with \$500 training cost by vectorized
  timestep adaptation.
\newblock \emph{arXiv preprint arXiv:2507.16116}, 2025{\natexlab{b}}.

\bibitem[Loshchilov \& Hutter(2017)Loshchilov and Hutter]{OPTIM:SGDR}
Ilya Loshchilov and Frank Hutter.
\newblock Sgdr: Stochastic gradient descent with warm restarts.
\newblock In \emph{ICLR}, 2017.

\bibitem[Loshchilov \& Hutter(2019)Loshchilov and Hutter]{OPTIM:ADAMW}
Ilya Loshchilov and Frank Hutter.
\newblock Decoupled weight decay regularization.
\newblock In \emph{ICLR}, 2019.

\bibitem[Ma et~al.(2025)Ma, Huang, Yan, Chen, Duan, Yin, Wan, Ming, Song, Chen,
  et~al.]{MODEL:STEPVIDEO}
Guoqing Ma, Haoyang Huang, Kun Yan, Liangyu Chen, Nan Duan, Shengming Yin,
  Changyi Wan, Ranchen Ming, Xiaoniu Song, Xing Chen, et~al.
\newblock Step-video-t2v technical report: The practice, challenges, and future
  of video foundation model.
\newblock \emph{arXiv preprint arXiv:2502.10248}, 2025.

\bibitem[Mentzer et~al.(2024)Mentzer, Minnen, Agustsson, and
  Tschannen]{TOKENIZER:FSQ}
Fabian Mentzer, David Minnen, Eirikur Agustsson, and Michael Tschannen.
\newblock Finite scalar quantization: {VQ}-{VAE} made simple.
\newblock In \emph{ICLR}, 2024.

\bibitem[Oord et~al.(2017)Oord, Vinyals, and Kavukcuoglu]{TOKENIZER:VQVAE}
Aaron van~den Oord, Oriol Vinyals, and Koray Kavukcuoglu.
\newblock Neural discrete representation learning.
\newblock In \emph{NeurIPS}, 2017.

\bibitem[Ou et~al.(2025)Ou, Nie, Xue, Zhu, Sun, Li, and Li]{MODEL:RADD}
Jingyang Ou, Shen Nie, Kaiwen Xue, Fengqi Zhu, Jiacheng Sun, Zhenguo Li, and
  Chongxuan Li.
\newblock Your absorbing discrete diffusion secretly models the conditional
  distributions of clean data.
\newblock In \emph{ICLR}, 2025.

\bibitem[Peng et~al.(2025)Peng, Zheng, Shen, Young, Guo, Wang, Xu, Liu, Jiang,
  Li, et~al.]{MODEL:OPENSORA}
Xiangyu Peng, Zangwei Zheng, Chenhui Shen, Tom Young, Xinying Guo, Binluo Wang,
  Hang Xu, Hongxin Liu, Mingyan Jiang, Wenjun Li, et~al.
\newblock Opensora 2.0: Training a commercial-level video generation model in
  \$200k.
\newblock \emph{arXiv preprint arXiv:2503.09642}, 2025.

\bibitem[Podell et~al.(2024)Podell, English, Lacey, Blattmann, Dockhorn,
  M{\"u}ller, Penna, and Rombach]{MODEL:SDXL}
Dustin Podell, Zion English, Kyle Lacey, Andreas Blattmann, Tim Dockhorn, Jonas
  M{\"u}ller, Joe Penna, and Robin Rombach.
\newblock Sdxl: Improving latent diffusion models for high-resolution image
  synthesis.
\newblock In \emph{ICLR}, 2024.

\bibitem[Radford et~al.(2018)Radford, Narasimhan, Salimans, Sutskever,
  et~al.]{LLM:GPT1}
Alec Radford, Karthik Narasimhan, Tim Salimans, Ilya Sutskever, et~al.
\newblock Improving language understanding by generative pre-training.
\newblock \emph{OpenAI Blog}, 2018.

\bibitem[Radford et~al.(2019)Radford, Wu, Child, Luan, Amodei, Sutskever,
  et~al.]{LLM:GPT2}
Alec Radford, Jeffrey Wu, Rewon Child, David Luan, Dario Amodei, Ilya
  Sutskever, et~al.
\newblock Language models are unsupervised multitask learners.
\newblock \emph{OpenAI Blog}, 2019.

\bibitem[Ramesh et~al.(2021)Ramesh, Pavlov, Goh, Gray, Voss, Radford, Chen, and
  Sutskever]{MODEL:DALLE1}
Aditya Ramesh, Mikhail Pavlov, Gabriel Goh, Scott Gray, Chelsea Voss, Alec
  Radford, Mark Chen, and Ilya Sutskever.
\newblock Zero-shot text-to-image generation.
\newblock In \emph{ICML}, 2021.

\bibitem[Reed et~al.(2017)Reed, Oord, Kalchbrenner, Colmenarejo, Wang, Chen,
  Belov, and Freitas]{MODEL:PARALLEL}
Scott Reed, A{\"a}ron Oord, Nal Kalchbrenner, Sergio~G{\'o}mez Colmenarejo,
  Ziyu Wang, Yutian Chen, Dan Belov, and Nando Freitas.
\newblock Parallel multiscale autoregressive density estimation.
\newblock In \emph{ICML}, 2017.

\bibitem[Ren et~al.(2024)Ren, Yang, Zhang, Wei, Du, Huang, and
  Chen]{MODEL:CONSISTI2V}
Weiming Ren, Huan Yang, Ge~Zhang, Cong Wei, Xinrun Du, Wenhao Huang, and Wenhu
  Chen.
\newblock Consisti2v: Enhancing visual consistency for image-to-video
  generation.
\newblock \emph{TMLR}, 2024.

\bibitem[Shaul et~al.(2025)Shaul, Gat, Havasi, Severo, Sriram, Holderrieth,
  Karrer, Lipman, and Chen]{ALGO:KINETIC}
Neta Shaul, Itai Gat, Marton Havasi, Daniel Severo, Anuroop Sriram, Peter
  Holderrieth, Brian Karrer, Yaron Lipman, and Ricky T.~Q. Chen.
\newblock Flow matching with general discrete paths: A kinetic-optimal
  perspective.
\newblock In \emph{ICLR}, 2025.

\bibitem[Shi et~al.(2025)Shi, Luo, Ge, Yang, Shan, and Wang]{TOKENIZER:IBQ}
Fengyuan Shi, Zhuoyan Luo, Yixiao Ge, Yujiu Yang, Ying Shan, and Limin Wang.
\newblock Scalable image tokenization with index backpropagation quantization.
\newblock In \emph{ICCV}, 2025.

\bibitem[Song et~al.(2021)Song, Sohl-Dickstein, Kingma, Kumar, Ermon, and
  Poole]{ALGO:SMLM}
Yang Song, Jascha Sohl-Dickstein, Diederik~P Kingma, Abhishek Kumar, Stefano
  Ermon, and Ben Poole.
\newblock Score-based generative modeling through stochastic differential
  equations.
\newblock In \emph{ICLR}, 2021.

\bibitem[Su et~al.(2024)Su, Ahmed, Lu, Pan, Bo, and Liu]{ARCH:ROPE}
Jianlin Su, Murtadha Ahmed, Yu~Lu, Shengfeng Pan, Wen Bo, and Yunfeng Liu.
\newblock Roformer: Enhanced transformer with rotary position embedding.
\newblock \emph{Neurocomputing}, 568:\penalty0 127063, 2024.

\bibitem[Sun et~al.(2023)Sun, Pan, Ge, Li, Duan, Wu, Zhang, Zhou, Qin, Wang,
  Dai, Qiao, Wang, and Li]{DATASET:JOURNEYDB}
Keqiang Sun, Junting Pan, Yuying Ge, Hao Li, Haodong Duan, Xiaoshi Wu, Renrui
  Zhang, Aojun Zhou, Zipeng Qin, Yi~Wang, Jifeng Dai, Yu~Qiao, Limin Wang, and
  Hongsheng Li.
\newblock Journey{DB}: A benchmark for generative image understanding.
\newblock In \emph{NeurIPS}, 2023.

\bibitem[Sun et~al.(2024{\natexlab{a}})Sun, Jiang, Chen, Zhang, Peng, Luo, and
  Yuan]{MODEL:LLAMAGEN}
Peize Sun, Yi~Jiang, Shoufa Chen, Shilong Zhang, Bingyue Peng, Ping Luo, and
  Zehuan Yuan.
\newblock Autoregressive model beats diffusion: Llama for scalable image
  generation.
\newblock \emph{arXiv preprint arXiv:2406.06525}, 2024{\natexlab{a}}.

\bibitem[Sun et~al.(2025)Sun, Jiang, Zhao, and He]{ALGO:NOISECOND-CFM}
Qiao Sun, Zhicheng Jiang, Hanhong Zhao, and Kaiming He.
\newblock Is noise conditioning necessary for denoising generative models?
\newblock In \emph{ICML}, 2025.

\bibitem[Sun et~al.(2024{\natexlab{b}})Sun, Cui, Zhang, Zhang, Yu, Wang, Rao,
  Liu, Huang, and Wang]{VLM:EMU2}
Quan Sun, Yufeng Cui, Xiaosong Zhang, Fan Zhang, Qiying Yu, Yueze Wang,
  Yongming Rao, Jingjing Liu, Tiejun Huang, and Xinlong Wang.
\newblock Generative multimodal models are in-context learners.
\newblock In \emph{CVPR}, 2024{\natexlab{b}}.

\bibitem[Tang et~al.(2025)Tang, Zheng, Paul, and Xie]{MODEL:FUSEDIT}
Bingda Tang, Boyang Zheng, Sayak Paul, and Saining Xie.
\newblock Exploring the deep fusion of large language models and diffusion
  transformers for text-to-image synthesis.
\newblock In \emph{CVPR}, 2025.

\bibitem[Tang et~al.(2022)Tang, Gu, Bao, Chen, and Wen]{ALGO:VQERROR}
Zhicong Tang, Shuyang Gu, Jianmin Bao, Dong Chen, and Fang Wen.
\newblock Improved vector quantized diffusion models.
\newblock \emph{arXiv preprint arXiv:2205.16007}, 2022.

\bibitem[Teng et~al.(2025)Teng, Jia, Sun, Li, Li, Tang, Han, Zhang, Zhang, Luo,
  et~al.]{MODEL:MAGI1}
Hansi Teng, Hongyu Jia, Lei Sun, Lingzhi Li, Maolin Li, Mingqiu Tang, Shuai
  Han, Tianning Zhang, WQ~Zhang, Weifeng Luo, et~al.
\newblock Magi-1: Autoregressive video generation at scale.
\newblock \emph{arXiv preprint arXiv:2505.13211}, 2025.

\bibitem[Unsplash(2020)]{DATASET:UNSPLASH}
Unsplash.
\newblock Unsplash dataset, 2020.

\bibitem[Wang et~al.(2025{\natexlab{a}})Wang, Ai, Wen, Mao, Xie, Chen, Yu,
  Zhao, Yang, et~al.]{MODEL:WAN}
Ang Wang, Baole Ai, Bin Wen, Chaojie Mao, Chen-Wei Xie, Di~Chen, Feiwu Yu,
  Haiming Zhao, Jianxiao Yang, et~al.
\newblock Wan: Open and advanced large-scale video generative models.
\newblock \emph{arXiv preprint arXiv:2503.20314}, 2025{\natexlab{a}}.

\bibitem[Wang et~al.(2025{\natexlab{b}})Wang, Lai, Li, Zhang, Sun, Kang, Wu,
  Li, and Luo]{MODEL:FUDOKI}
Jin Wang, Yao Lai, Aoxue Li, Shifeng Zhang, Jiacheng Sun, Ning Kang, Chengyue
  Wu, Zhenguo Li, and Ping Luo.
\newblock Fudoki: Discrete flow-based unified understanding and generation via
  kinetic-optimal velocities.
\newblock In \emph{NeurIPS}, 2025{\natexlab{b}}.

\bibitem[Wang et~al.(2024{\natexlab{a}})Wang, Bai, Tan, Wang, Fan, Bai, Chen,
  Liu, Wang, Ge, et~al.]{VLM:QWEN2VL}
Peng Wang, Shuai Bai, Sinan Tan, Shijie Wang, Zhihao Fan, Jinze Bai, Keqin
  Chen, Xuejing Liu, Jialin Wang, Wenbin Ge, et~al.
\newblock Qwen2-vl: Enhancing vision-language model's perception of the world
  at any resolution.
\newblock \emph{arXiv preprint arXiv:2409.12191}, 2024{\natexlab{a}}.

\bibitem[Wang et~al.(2025{\natexlab{c}})Wang, Shi, Ou, Chen, Lin, Wang, Jiang,
  Yang, Zheng, Tao, et~al.]{DATASET:KOALA}
Qiuheng Wang, Yukai Shi, Jiarong Ou, Rui Chen, Ke~Lin, Jiahao Wang, Boyuan
  Jiang, Haotian Yang, Mingwu Zheng, Xin Tao, et~al.
\newblock Koala-36m: A large-scale video dataset improving consistency between
  fine-grained conditions and video content.
\newblock In \emph{CVPR}, 2025{\natexlab{c}}.

\bibitem[Wang et~al.(2024{\natexlab{b}})Wang, Zhang, Luo, Sun, Cui, Wang,
  Zhang, Wang, Li, Yu, et~al.]{VLM:EMU3}
Xinlong Wang, Xiaosong Zhang, Zhengxiong Luo, Quan Sun, Yufeng Cui, Jinsheng
  Wang, Fan Zhang, Yueze Wang, Zhen Li, Qiying Yu, et~al.
\newblock Emu3: Next-token prediction is all you need.
\newblock \emph{arXiv preprint arXiv:2409.18869}, 2024{\natexlab{b}}.

\bibitem[Wang et~al.(2024{\natexlab{c}})Wang, Xiong, Zhou, Lin, Zhao, Kang,
  Feng, and Liu]{MODEL:LOONG}
Yuqing Wang, Tianwei Xiong, Daquan Zhou, Zhijie Lin, Yang Zhao, Bingyi Kang,
  Jiashi Feng, and Xihui Liu.
\newblock Loong: Generating minute-level long videos with autoregressive
  language models.
\newblock \emph{arXiv preprint arXiv:2410.02757}, 2024{\natexlab{c}}.

\bibitem[Wu et~al.(2025{\natexlab{a}})Wu, Li, Zhou, Lin, Gao, Yan, Yin, Bai,
  Xu, Chen, et~al.]{MODEL:QWENIMAGE}
Chenfei Wu, Jiahao Li, Jingren Zhou, Junyang Lin, Kaiyuan Gao, Kun Yan,
  Sheng-ming Yin, Shuai Bai, Xiao Xu, Yilei Chen, et~al.
\newblock Qwen-image technical report.
\newblock \emph{arXiv preprint arXiv:2508.02324}, 2025{\natexlab{a}}.

\bibitem[Wu et~al.(2025{\natexlab{b}})Wu, Zheng, Yan, Xiao, Luo, Wang, Li,
  Jiang, Liu, Zhou, et~al.]{MODEL:OMNIGEN2}
Chenyuan Wu, Pengfei Zheng, Ruiran Yan, Shitao Xiao, Xin Luo, Yueze Wang, Wanli
  Li, Xiyan Jiang, Yexin Liu, Junjie Zhou, et~al.
\newblock Omnigen2: Exploration to advanced multimodal generation.
\newblock \emph{arXiv preprint arXiv:2506.18871}, 2025{\natexlab{b}}.

\bibitem[Xie et~al.(2025{\natexlab{a}})Xie, Chen, Zhao, YU, Zhu, Lin, Zhang,
  Li, Chen, Cai, Liu, Zhou, and Han]{MODEL:SANA1.5}
Enze Xie, Junsong Chen, Yuyang Zhao, Jincheng YU, Ligeng Zhu, Yujun Lin, Zhekai
  Zhang, Muyang Li, Junyu Chen, Han Cai, Bingchen Liu, Daquan Zhou, and Song
  Han.
\newblock Sana 1.5: Efficient scaling of training-time and inference-time
  compute in linear diffusion transformer.
\newblock In \emph{ICML}, 2025{\natexlab{a}}.

\bibitem[Xie et~al.(2025{\natexlab{b}})Xie, Darrell, Zettlemoyer, and
  Wang]{MODEL:RECA}
Ji~Xie, Trevor Darrell, Luke Zettlemoyer, and XuDong Wang.
\newblock Reconstruction alignment improves unified multimodal models.
\newblock \emph{arXiv preprint arXiv:2509.07295}, 2025{\natexlab{b}}.

\bibitem[Xie et~al.(2025{\natexlab{c}})Xie, Mao, Bai, Zhang, Wang, Lin, Gu,
  Chen, Yang, and Shou]{MODEL:SHOWO}
Jinheng Xie, Weijia Mao, Zechen Bai, David~Junhao Zhang, Weihao Wang,
  Kevin~Qinghong Lin, Yuchao Gu, Zhijie Chen, Zhenheng Yang, and Mike~Zheng
  Shou.
\newblock Show-o: One single transformer to unify multimodal understanding and
  generation.
\newblock In \emph{ICLR}, 2025{\natexlab{c}}.

\bibitem[Xie et~al.(2025{\natexlab{d}})Xie, Yang, and Shou]{MODEL:SHOWO2}
Jinheng Xie, Zhenheng Yang, and Mike~Zheng Shou.
\newblock Show-o2: Improved native unified multimodal models.
\newblock In \emph{NeurIPS}, 2025{\natexlab{d}}.

\bibitem[Xing et~al.(2024)Xing, Xia, Zhang, Chen, Yu, Liu, Liu, Wang, Shan, and
  Wong]{MODEL:DYNAMICRAFTER}
Jinbo Xing, Menghan Xia, Yong Zhang, Hao Chen, Wangbo Yu, Hanyuan Liu, Gongye
  Liu, Xintao Wang, Ying Shan, and Tien-Tsin Wong.
\newblock Dynamicrafter: Animating open-domain images with video diffusion
  priors.
\newblock In \emph{ECCV}, 2024.

\bibitem[Yan et~al.(2021)Yan, Zhang, Abbeel, and Srinivas]{MODEL:VIDEOGPT}
Wilson Yan, Yunzhi Zhang, Pieter Abbeel, and Aravind Srinivas.
\newblock Videogpt: Video generation using vq-vae and transformers.
\newblock \emph{arXiv preprint arXiv:2104.10157}, 2021.

\bibitem[Yang et~al.(2025{\natexlab{a}})Yang, Li, Yang, Zhang, Hui, Zheng, Yu,
  Gao, Huang, Lv, et~al.]{LLM:QWEN3}
An~Yang, Anfeng Li, Baosong Yang, Beichen Zhang, Binyuan Hui, Bo~Zheng, Bowen
  Yu, Chang Gao, Chengen Huang, Chenxu Lv, et~al.
\newblock Qwen3 technical report.
\newblock \emph{arXiv preprint arXiv:2505.09388}, 2025{\natexlab{a}}.

\bibitem[Yang et~al.(2025{\natexlab{b}})Yang, Teng, Zheng, Ding, Huang, Xu,
  Yang, Hong, Zhang, Feng, Yin, Yuxuan.Zhang, Wang, Cheng, Xu, Gu, Dong, and
  Tang]{MODEL:COGVIDEOX}
Zhuoyi Yang, Jiayan Teng, Wendi Zheng, Ming Ding, Shiyu Huang, Jiazheng Xu,
  Yuanming Yang, Wenyi Hong, Xiaohan Zhang, Guanyu Feng, Da~Yin, Yuxuan.Zhang,
  Weihan Wang, Yean Cheng, Bin Xu, Xiaotao Gu, Yuxiao Dong, and Jie Tang.
\newblock Cogvideox: Text-to-video diffusion models with an expert transformer.
\newblock In \emph{ICLR}, 2025{\natexlab{b}}.

\bibitem[Yu et~al.(2025)Yu, Gong, Yuan, Zheng, Chai, Chen, Zheng, and
  Zhao]{MODEL:VIDEOMAR}
Hu~Yu, Biao Gong, Hangjie Yuan, DanDan Zheng, Weilong Chai, Jingdong Chen,
  Kecheng Zheng, and Feng Zhao.
\newblock Videomar: Autoregressive video generatio with continuous tokens.
\newblock In \emph{NeurIPS}, 2025.

\bibitem[Yu et~al.(2022)Yu, Xu, Koh, Luong, Baid, Wang, Vasudevan, Ku, Yang,
  Ayan, Hutchinson, Han, Parekh, Li, Zhang, Baldridge, and Wu]{MODEL:PARTI}
Jiahui Yu, Yuanzhong Xu, Jing~Yu Koh, Thang Luong, Gunjan Baid, Zirui Wang,
  Vijay Vasudevan, Alexander Ku, Yinfei Yang, Burcu~Karagol Ayan, Ben
  Hutchinson, Wei Han, Zarana Parekh, Xin Li, Han Zhang, Jason Baldridge, and
  Yonghui Wu.
\newblock Scaling autoregressive models for content-rich text-to-image
  generation.
\newblock \emph{TMLR}, 2022.

\bibitem[Yu et~al.(2023)Yu, Cheng, Sohn, Lezama, Zhang, Chang, Hauptmann, Yang,
  Hao, Essa, et~al.]{MODEL:MAGVIT}
Lijun Yu, Yong Cheng, Kihyuk Sohn, Jos{\'e} Lezama, Han Zhang, Huiwen Chang,
  Alexander~G Hauptmann, Ming-Hsuan Yang, Yuan Hao, Irfan Essa, et~al.
\newblock Magvit: Masked generative video transformer.
\newblock In \emph{CVPR}, 2023.

\bibitem[Yuan et~al.(2025)Yuan, Chen, Cen, Yu, Liang, Chang, Lin, Feng, Liu,
  Xing, et~al.]{MODEL:LUMOS}
Hangjie Yuan, Weihua Chen, Jun Cen, Hu~Yu, Jingyun Liang, Shuning Chang, Zhihui
  Lin, Tao Feng, Pengwei Liu, Jiazheng Xing, et~al.
\newblock Lumos-1: On autoregressive video generation from a unified model
  perspective.
\newblock \emph{arXiv preprint arXiv:2507.08801}, 2025.

\bibitem[Zheng et~al.(2025)Zheng, Chen, Mao, Liu, Zhu, and
  Zhang]{ALGO:NOISECOND-MDM}
Kaiwen Zheng, Yongxin Chen, Hanzi Mao, Ming-Yu Liu, Jun Zhu, and Qinsheng
  Zhang.
\newblock Masked diffusion models are secretly time-agnostic masked models and
  exploit inaccurate categorical sampling.
\newblock In \emph{ICLR}, 2025.

\bibitem[Zhu et~al.(2025)Zhu, Wang, Lathuili\`ere, and Kalogeiton]{ALGO:Dimo}
Yuanzhi Zhu, Xi~Wang, St\'ephane Lathuili\`ere, and Vicky Kalogeiton.
\newblock Di[m]o: Distilling masked diffusion models into one-step generator.
\newblock In \emph{ICCV}, 2025.

\end{thebibliography}
\bibliographystyle{reference}

\clearpage
\appendix
\section*{Appendix}

We have published code and pre-trained models to improve interpretability and ensure reproducibility. 
In this appendix, implementation details, experiments, and qualitative results are organized as follows:

{\leftmargini=2em
\begin{itemize}
\item Training and Sampling Details (Sec.~\ref{sec:supp:training_sampling})
\item Video extrapolation experiments (Sec.~\ref{sec:supp:extrapolation})
\item Start-End frame control experiments (Sec.~\ref{sec:supp:startend})
\end{itemize}}

\section{Training and Sampling Details}\label{sec:supp:training_sampling}
\vspace{-2em}
\noindent
\begin{minipage}[t]{0.48\linewidth}
\begin{algorithm}[H]
\caption{\Ours Training}\label{alg:train}
\begin{algorithmic}[1]
\Require Predictor $p_\theta$, Shift $\lambda$
\Repeat
    \State $x_1 \sim p_{\text{data}}$
    \State $t \sim \mathcal{U}(0,1)$
    \State $\tilde{t} \gets t / (t + \lambda(1 - t))$
    \Comment{time shift}
    \State $x_{\tilde{t}} \sim p_{\tilde{t}|1}(\cdot|x_1)$
    \State $\mathcal{L} \gets -\sum_{i=1}^{D} \log p_\theta(x_1^i \mid x_{\tilde{t}})$
    \State $\theta \gets \theta - \eta \nabla_\theta \mathcal{L}$
\Until{converged}
\State \Return Trained predictor $p_\theta$
\end{algorithmic}
\end{algorithm}
\end{minipage}
\hfill
\begin{minipage}[t]{0.48\linewidth}
\begin{algorithm}[H]
\caption{\Ours Sampling}\label{alg:infer}
\begin{algorithmic}[1]
\Require Predictor $p_\theta$, Steps $T$, Shift $\lambda$, Visual vocabulary $\mathcal{V}$
    \State Sample $x_0 \sim \mathrm{Uniform}(\mathcal{V})$
    \For{$k = 1$ to $T$}
        \State $t \gets (k-1)/T$
        \State $\tilde{t} \gets t / (t + \lambda(1 - t))$
        \Comment{time shift}
        \State $\hat{x}_1 \sim p_\theta(\cdot | x_{\tilde{t}})$
        \State {$u_{\tilde{t}} \gets u_t(x, z | \hat{x}_1)$}
        \State $x_{\tilde{t}} \gets x_{\tilde{t}} + h \cdot u_{\tilde{t}}$
    \EndFor
    \State \Return $x_1$ \Comment{Generated discrete sample}
\end{algorithmic}
\end{algorithm}
\end{minipage}

\vspace{1em}
\textbf{Training.} \OursRegular is a predictor, a parametric model $p_{\theta}(\cdot|x_t)$, that takes $x_{t}$ as input and refine all tokens simultaneously.We first encode images and videos into discrete latent tokens using a pre-trained tokenizer. For visual tokens, we adopt a DFM training objective based on the probability path. At each iteration, we randomly sample a timestep $t \in [0, 1]$ and use the metric path to obtain the noised tokens $x_t$. Text prompts are tokenized using the Qwen3 tokenizer and embedded into the same semantic space. We concatenate text embeddings and noised visual tokens into a unified sequence. The training objective is defined as the expected cross-entropy between the ground-truth visual token sequence and the model's predicted distribution. We list the detailed training process in Algorithm~\ref{alg:train}.

\textbf{Sampling.} This velocity field ensures that transitions occur only from state $z$ to state $x$ when $x$ is closer to $x_1$ than $z$, \ie, $d(x, x_1) < d(z, x_1)$. Using the distance metric and the time-dependent factor $\beta_t$, the velocity guides the flow of particles in a manner that is both kinetic-optimal and aligned with the underlying geometry of the state space. We list the complete sampling process in Algorithm~\ref{alg:infer}. 

\section{Video extrapolation experiments}\label{sec:supp:extrapolation}
\begin{figure}[ht!]
\vspace{-3em}
\centering
\includegraphics[width=0.97\linewidth]{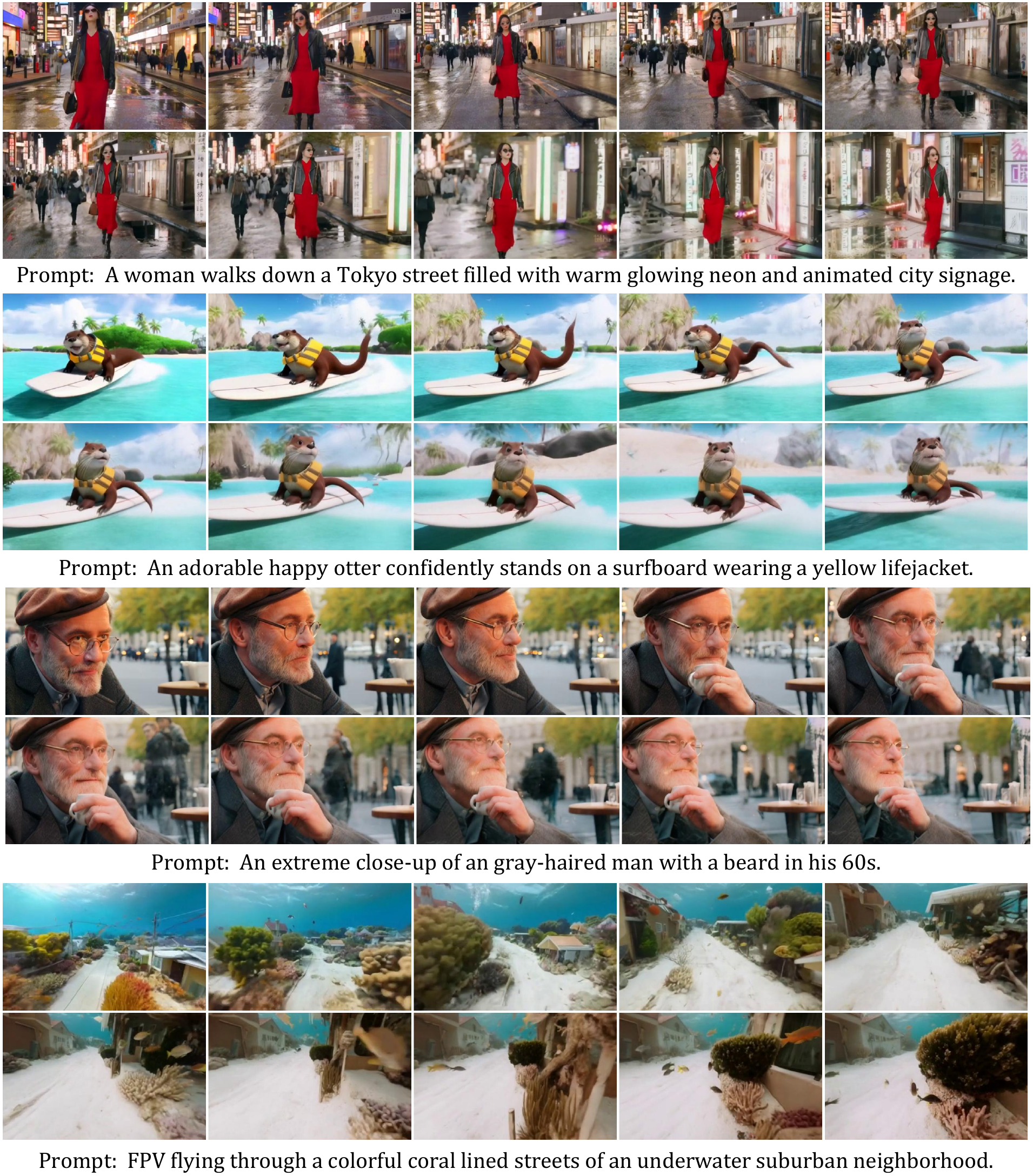}
\caption{\textbf{Zero-shot video extrapolation.} We extend the 4-second text-to-video result to 40 seconds.}
\label{fig:supp:extrapolation}
\end{figure}

As \OursRegular is trained by applying independent noise levels to each frame,
it naturally lends itself to video extrapolation via a sliding window.
Specifically, new frames are generated sequentially, conditioned on the most recent 13 frames, thereby extending future predictions beyond the initial 49-frame context window.
To effectively mitigate sampling errors in autoregressive video generation, we introduce a small amount of noise into historical frames by resampling them at timestep $t=0.9$.
Figure~\ref{fig:supp:extrapolation} presents the qualitative results for a video of 481 frames, where the initial text-to-video segment is extended through 12 extrapolation steps, producing videos up to 10$\times$ the original length.

\section{Start-End frame control experiments}\label{sec:supp:startend}
\begin{figure}[ht!]
\centering
\includegraphics[width=0.97\linewidth]{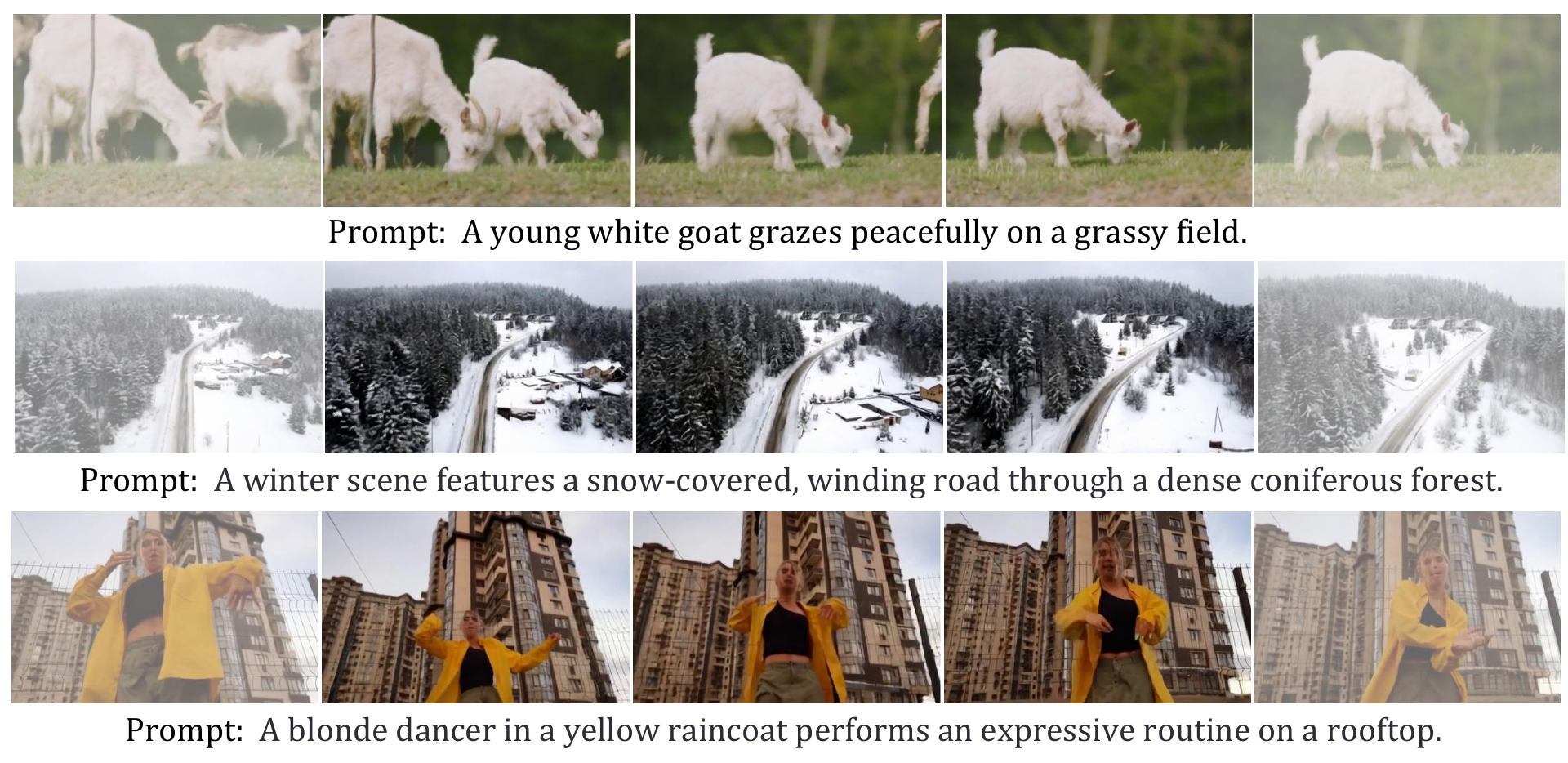}
\caption{\textbf{Zero-shot start-end frame control.} The start-end frames are rendered with transparency.}
\label{fig:supp:startend}
\end{figure}

We evaluate \OursRegular on the start-end frame control task, a specialized form of video generation to prevent future predictions from drifting.
Concretely, we extract a sequence of frames from the video at 4-second intervals and place them sequentially at the beginning and the end of the context window.
This setup enables the generation of a video featuring coherent motion of both objects and cameras, preserving spatial relationships throughout the scene.
We present the qualitative results in Figure~\ref{fig:supp:startend}.

\end{document}